\documentclass{article}
\usepackage{booktabs}
\usepackage{fontspec}
\usepackage[T1]{fontenc}
\usepackage{rotating}
\usepackage{float}
\usepackage{xcolor}
\usepackage{bm}
\usepackage{kbordermatrix}
\usepackage{fullpage}
\usepackage{url}
\usepackage[english]{babel}
\usepackage{graphicx}
\usepackage{color}
\usepackage{rotating}
\usepackage{tabularx}
\usepackage[citecolor=blue,colorlinks=true,filecolor=blue,urlcolor=blue]{hyperref}
\usepackage{natbib}
\usepackage{tipa}
\usepackage{amsmath}
\usepackage{comment}
\usepackage{authblk}
\title{Analyzing Finnish Inflectional Classes through Discriminative Lexicon and Deep Learning Models}

\author[1]{Alexandre Nikolaev}
\author[2]{Yu-Ying Chuang}
\author[3]{R. Harald Baayen}
\affil[1]{University of Eastern Finland}
\affil[2]{National Taiwan Normal University}
\affil[3]{University of T{\"u}bingen}

\date{}

\begin{document}

\maketitle

\begin{abstract}

\noindent
Descriptions of complex nominal or verbal systems make use of inflectional classes. Inflectional classes bring together nouns which have similar stem changes and use similar exponents in their paradigms. Establishing what the inflectional classes of a language are is far from trivial, and across grammatical descriptions, the number of classes distinguished can vary considerably. Although inflectional classes can be very useful for language teaching as well as for setting up finite state morphological systems, it is unclear whether inflectional classes are cognitively real, in the sense that native speakers would need to discover these classes in order to learn how to properly inflect the nouns of their language. This study investigates whether the Discriminative Lexicon Model (DLM) can understand and produce Finnish inflected nouns without setting up inflectional classes, using a dataset with 55,271 inflected nouns of 2000 high-frequency Finnish nouns from 49 inflectional classes.  
%
Several DLM comprehension and production models were set up. Some models were not informed about frequency of use, and provide insight into learnability with infinite exposure (endstate learning). Other models were set up from a usage based perspective, and were trained with token frequencies being taken into consideration (frequency-informed learning). On training data, models performed with very high accuracies. For held-out test data, accuracies decreased, as expected, but remained acceptable. Across most models, performance increased for inflectional classes with more types, more lower-frequency words, and more hapax legomena,  mirroring the productivity of the inflectional classes. The model struggles more with novel forms of unproductive and less productive classes, and performs far better for unseen forms belonging to productive classes.  However, for usage-based production models, frequency was the dominant predictor of model performance, and correlations with measures of productivity were tenuous or absent.  As usage-based comprehension models show strong correlations with the productivity measures, our results indicate a novel asymmetry between comprehension and production.  Our study also demonstrates that the inflectional system of Finnish nouns can be learned surprisingly well without hand-crafting of inflectional classes, challenging the assumption that explicit inflectional classes features have to be mastered by language learners.     
\end{abstract}


\section{Introduction}

\noindent
Descriptions of complex nominal or verbal inflectional systems make use of inflectional classes. Nominal inflectional classes bring together nouns which have similar stem changes and use similar exponents in their paradigms. Establishing what the inflectional classes of a language are is far from trivial, and across grammatical descriptions, the number of classes distinguished can vary considerably. Although inflectional classes can be very useful for language teaching as well as for setting up finite state morphological systems, it is unclear whether inflectional classes are cognitively real, in the sense that native speakers would need to discover these classes in order to learn how to properly inflect the nouns of their language. \textcolor{black}{This skepticism aligns with findings suggesting that even the concept of a full paradigm is largely an analytical construct, with learners typically encountering only sparse, overlapping subsets of word forms \citep{janda2021less}.} This study investigates whether the Discriminative Lexicon Model (DLM), a cognitively motivated computational model of the mental lexicon, can understand and produce Finnish inflected nouns without setting up explicit inflectional classes.  Since computational experiments with the DLM support this possibility, a second question that we address is why it is possible for the DLM to master Finnish inflection without explicit knowledge of inflectional classes.  We will show that the inflectional classes are to some extent reflected in Finnish word embeddings, 
and argue that it is this kind of isomorphy between form and meaning that likely enables learning in the DLM.
We will also show that how well the DLM is able to understand and produce novel forms is correlated with the degrees of productivity of the different inflectional classes.  

The present study takes a usage-based approach \citep{bybee1995regular} to morphological productivity, and therefore considers computational models that take frequency of use into account \textcolor{black}{(see also \cite{janda2021less} as an example of research experimentally validating the effectiveness of learning from frequency-driven, partial input, as opposed to abstract, complete structures)}. We compare these models with corresponding models from which frequency is withheld.  Such models provide insight into what can be learned with infinite experience and shed light on pure type-based systematicities.  Conversely, the models that take frequency into account approximate much better the constraints under which human learning takes place. 

Our approach differs markedly from the proposal in generative grammar that productivity would be governed by a mechanism that allows learners to maintain productive rules only as long as the burden of exceptions remains below a quantifiable threshold \citep[the tolerance principle]{yang2016price}.  According to this principle, morphological rules are either fully productive, or fully unproductive. Furthermore, this rule vs. rote approach requires the analyst to evaluate whether a word is exceptional.  For Finnish inflectional classes, this is an impossible task. We shall see that setting up inflectional classes for Finnish is far from trivial, and that it is unclear how to differentiate between exceptions and subregularities. Furthermore, we will also show that at the level of word meaning, rigid distinctions between regular and irregular semantics cannot be made.   The computational modeling experiments that we report below provide evidence that the nominal morphology of Finnish nouns can be learned without symbolic abstractions, by exploiting the statistical regularities that govern the relation between form and meaning in Finnish,
very much along the lines envisioned 30 years ago by \cite{bybee1995regular}. \textcolor{black}{Similarly, recent work by \cite{divjak2025learnability} used computational modeling based on error-correction learning to investigate whether the abstract category of grammatical aspect in Polish is necessary for predicting usage, finding that models relying on concrete lexical and contextual cues better captured user behavior than those based on abstract semantic definitions.}

In the remainder of this study, we first introduce the intricacies of the inflectional classes of Finnish. Subsequently, we present an analysis of the Finnish embeddings, using linear discriminant analysis to probe to what extent inflectional class structure is mirrored in semantic space. We then introduce the DLM, and report  computational experiments addressing the productivity of the inflectional classes for both comprehension and production.   The paper concludes with a general discussion.

\section{Finnish inflectional classes}

\subsection{Historical background}

\noindent
The first published (in 1649) Finnish grammar \citep{petraeus19681649}  was written in Latin. It largely copied the model of Latin grammar, which Petræus took from the then-fashionable Latin grammars (especially Tiderus, Chytrus, Donatus, and Melanchthon) \citep{vihonen1978}. Following common 17th-century practise, Petræus's grammar provides a contrastive analysis that presents the morphological and semantic equivalents of Latin forms in Finnish \citep{vihonen1978suomen}. According to Petræus's grammar, Finnish nouns belong to eight different declension classes, with membership being based on the vowel preceding the \textit{n} in the genitive singular (a, e, i, o, u, ä, ö, y). From the perspective of a modern Finnish speaker's intuition, such a declension division may seem strange. However, for example, in the comprehensive grammar of Finnish \citep{vilkuna2004iso}, nouns are first divided into four inflectional types and then further sub-divided based exactly on the final vowel of the stem.  Whereas in current morphological analyses, Finnish case endings are taken to be exponents that do not vary much between inflectional classes, in Petræus's grammar, following Latin, each declension is analysed as having its own case endings. E.g., Petræus identified eight distinct genitive endings: -an, -en, -in, -on, -un, -än, -ön, -yn, treating them as separate suffixes, mirroring how Latin nouns have distinct terminations per declension (e.g., -us, -um, -ae) \citep{wiik1989suomen}. According to modern Finnish linguistics, the preceding vowels are part of the stem or adjusted for harmony, not inherent to the genitive suffix (-n) itself.

The adherence of the first grammarians to Latin grammar models is explained by the general view at the time that the categories of Latin morphology were reflections of natural concepts and modes of thought \citep{karlsson1997yleinen}. Later, this view crystallized in the form of a posited general grammar (Grammaire générale), according to which any given language is a copy or image of the structure of the world. As a consequence, this general grammar can be used in the study of any language. The hypothesis of a general grammar deduced from Latin and also French was highly influential in Finland in the 19th century, and dominated the grammars of the 1850s and 1860s, but lost influence in the early 20th century \citep{karlsson1998vuosien, karlsson2000setala}.


Starting from Vhael (1733, \citealp{vhael19681733}), grammarians noticed a key difference in how Finnish and Latin handle inflection. In Finnish, a single affix--known as an exponent--can be used across all noun types, regardless of their declension class. Unlike Latin, where endings vary depending on the declension class, Finnish relies on changes in the stem to define how a noun inflects. This contrast prompted several linguists, beginning with \cite{wikstrom1832forsok}, to abandon the use of declension classes entirely when describing Finnish grammar.

This shift, however, sparked a new debate: how should the basic form of a Finnish noun be defined? Some linguists, such as \cite{setala1898suomen}, criticized the practice of using a stem as the basic form. He argued that this approach was overly rigid, borrowed from the grammars of other languages, and failed to reflect the true nature of Finnish. In contrast, \cite{lonnrot1841}  and later \cite{wiik1967suomen} proposed using the strong vowel stem as the basic form. For example, for the noun \textit{käsi} `hand', they suggested \textit{käte-} as the basic form because it underlies many inflected forms, such as \textit{käteen} (used in the illative case, meaning `into the hand'). On the other hand, \cite{karlsson1977eraista, karlsson1983suomen} argued that the nominative singular, such as \textit{käsi}, should serve as the basic form.

%
%
%
%
%

Current analyses of Finnish nominal paradigms rely on two key elements: stems and exponents. These analyses reveal systematic relationships not only among the exponents but also among the various stems a noun may have \citep{karlsson1985paradigms}. Finnish nouns often exhibit multiple stem forms, a phenomenon stemming from historical sound changes that introduced alterations to the stem. These alterations can be either phonologically conditioned, meaning they are still governed by active phonological rules, or morphologically conditioned, indicating they are historical and no longer productive. Morphologically conditioned stem variations, while predictable, require additional knowledge of the noun's inflectional class. For example, if a noun's nominative form ends in \textit{-i} and belongs to a specific inflectional class, one can predict that its other stems will incorporate \textit{te-} or \textit{de-}, such as \textit{käte-} or \textit{käde-} depending on the following exponent. 

In current lexicography, inflectional classes are used to group similar paradigms together.
%
%
In the 20th century, two extensive monolingual Finnish dictionaries were published: the Modern Finnish Dictionary (MFD, \citealp{nykysuomen_sanakirja}) and the Basic Dictionary of Finnish (DoF, \citealp{haarala1990}). Both use a similar notation system, where each noun is accompanied by a number referring to a comprehensive table of inflectional classes.  Additionally, in MFD, nouns subject to gradation are marked with an asterisk, and in DoF, they are marked with their own code, which is explained in a separate gradation table. In Finnish consonant gradation, certain consonants (particularly p, t, and k) alternate between “strong” and “weak” forms depending on the morphological context (such as when adding suffixes). Quantitative gradation involves a change in the length of the consonant, often a geminate consonant (e.g., pp, tt, kk) shortens to a single consonant (p, t, k; e.g., \textit{kukka} `flower nom.sg.' : \textit{kukan} `flower gen.sg.'). Qualitative gradation involves a change in the quality of the consonant itself, such as k becoming a glottal stop or disappearing (e.g., \textit{ruoka} `food nom.sg.', \textit{ruoan} `food gen.sg.'). Table \ref{tab:four-paradigms} illustrates how three Finnish nouns (\textit{mato} `worm', \textit{matto} `carpet', and \textit{masto} `mast') inflect across all the Finnish noun cases. Even though these nouns differ in their consonant gradation patterns---\textit{mato} undergoing qualitative gradation (t : d), \textit{matto} featuring quantitative gradation (geminate tt : t), and \textit{masto} showing no gradation---they are classified in the same inflectional type (\#1) according to DoF. In other words, they share the same overall pattern of case endings, despite the variation in how their stems respond to suffixation. We also included the noun \textit{made} `burbot (a type of fish)' as an example of another inflectional type (\#48), to illustrate contrasts in stem alternation and case marking patterns.

The inflection type classification in MFD is the most extensive, comprising 82 nominal inflectional classes. This is not coincidental, as MFD is the most comprehensive general language dictionary of Finnish, and its table of inflection types is intended to serve practical purposes, not just to be evaluated from a theoretical perspective \citep{karlsson1983suomen}. Furthermore, MFD includes a large number of words that are no longer used in contemporary language.  \cite{karlsson1977eraista} criticizes MFD as an application of the WP model (Word and Paradigm), where the number of word paradigms is quite large, as even minor morphophonological differences can be the basis for paradigmatic differentiation. He gives an example of inflection types \#40 \textit{susi} `wolf' and \#41 \textit{tosi} `true,' whose only difference is the marginal plural genitive form \textit{sutten} in addition to \textit{susien}, cf. *\textit{totten}, \textit{tosien}.

The inflection type table in the DoF that was published later is more concise. In DoF, e.g., \textit{susi} and \textit{tosi} belong to the same 27th inflectional type. Similar types have been combined (\textit{maa} `earth / ground / land', \textit{pii} `silicon', \textit{puu} `tree'), obsolete ones removed (\textit{kiiru}) `haste', and some added (\textit{solakka} `slim'). The aim has been to show the usual inflection pattern of Finnish words according to the current standard language: clear dialectal and poetic forms have been eliminated. Thus, DoF does not mention inflectional forms such as \textit{vapahan} gen.sg. of `free' or \textit{vanhempata} part.sg. of `parent'. However, some forms that have become rare have been retained; their rarity is indicated in the table with parentheses (e.g., sg.illative form \textit{vieraihin} `guest' or pl.genitive form \textit{tuhanten} `thousand'). Some inflection forms marked as consonant-stemmed in MFD (\textit{suksi}:\textit{susta} `ski', \textit{lahti}:\textit{lahta} `bay') are only vowel-stemmed in DoF (\textit{suksea}, \textit{lahtea}) \citep{eronen1994}.

Since the noun paradigm classification system provided by DoF is currently the most comprehensive and modern, it is used in the present study with its vocabulary divided into 49 nominal inflectional classes.  It is worth keeping in mind, however, that this classification is far from perfect, and strikes a balance between generalization and exception listing that is currently perceived as an elegant compromise.  

\begin{table}[ht]
\centering
\begin{tabular}{lcccccccc}
\toprule
\textbf{Case} &
\multicolumn{2}{c}{\textbf{mato} `worm'} &
\multicolumn{2}{c}{\textbf{matto} `carpet'} &
\multicolumn{2}{c}{\textbf{masto} `mast'} &
\multicolumn{2}{c}{\textbf{made} `burbot'} \\
\cmidrule(lr){2-3}\cmidrule(lr){4-5}\cmidrule(lr){6-7}\cmidrule(lr){8-9}
 & \textbf{Sg} & \textbf{Pl} 
 & \textbf{Sg} & \textbf{Pl} 
 & \textbf{Sg} & \textbf{Pl} 
 & \textbf{Sg} & \textbf{Pl} \\
\midrule
Nominative   & mato    & madot       & matto    & matot       & masto     & mastot       & made     & mateet \\
Genitive     & madon   & matojen     & maton    & mattojen    & maston    & mastojen     & mateen   & mateiden/\\
&&&&&&&&mateitten\\
Partitive    & matoa   & matoja      & mattoa   & mattoja     & mastoa    & mastoja      & madetta  & mateita \\
Illative     & matoon  & matoihin    & mattoon  & mattoihin   & mastoon   & mastoihin    & mateeseen & mateihin/ \\
&&&&&&&&mateisiin\\
Inessive     & madossa & madoissa    & matossa  & matoissa    & mastossa  & mastoissa    & mateessa  & mateissa \\
Elative      & madosta & madoista    & matosta  & matoista    & mastosta  & mastoista    & mateesta  & mateista \\
Allative     & madolle & madoille    & matolle  & matoille    & mastolle  & mastoille    & mateelle  & mateille \\
Adessive     & madolla & madoilla    & matolla  & matoilla    & mastolla  & mastoilla    & mateella  & mateilla \\
Ablative     & madolta & madoilta    & matolta  & matoilta    & mastolta  & mastoilta    & mateelta  & mateilta \\
Essive       & matona  & matoina     & mattona  & mattoina    & mastona   & mastoina     & mateena   & mateina \\
Translative  & madoksi & madoiksi    & matoksi  & matoiksi    & mastoksi  & mastoiksi    & mateeksi  & mateiksi \\
Abessive     & madotta & matoitta    & matotta  & matoitta    & mastotta  & mastoitta    & mateetta  & mateitta \\
Instructive  & –       & madoin      & –        & matoin      & –         & mastoin      &     & matein \\
Comitative   & –       & matoine-    & –        & mattoine-   & –         & mastoine-    &   & mateine- \\
\bottomrule
\end{tabular}
\caption{Inflection of three Finnish nouns (\textit{mato} `worm', \textit{matto} `carpet', and \textit{masto} `mast') from inflectional type \#1 (DoF) and one noun (\textit{made} `burbot') from inflectional type \#48 (DoF) across all noun cases in singular (sg) and plural (pl).}
\label{tab:four-paradigms}
\end{table}

These complexities (see Table \ref{tab:four-paradigms}), where multiple paradigms---such as \textit{mato}, \textit{matto}, and \textit{masto}---exhibit distinct consonant gradation patterns yet belong to the same inflectional type (\#1, DoF), differing in stems and certain exponents (e.g., illative singular) from another type (\#48, DoF, \textit{made}), must be mastered by the DLM to effectively generalize across shared morphological structures.

\subsection{Productivity of inflectional types}

Inflectional classes typically differ with respect to the number of lexemes that they comprise.  They also differ with respect to their productivity.  From a diachronic perspective, new words that enter the language will typically be absorbed by one of the productive classes.  From a synchronic perspective, the question arises for existing lemmas whether low-frequency inflectional forms that language users have not encountered before will follow the systematicities of the inflectional class to which these lemmas belong.  As for lower-frequency nouns many inflectional variants are not attested even in large corpora, but can be readily understood and created when necessary \citep{karlsson1986frequency,Loo:2018b}. This makes the question far from trivial.

Class size and productivity are not the same. As observed by \citet{Anshen:Aronoff:81}, ``Apparently, a language can have in its lexicon a fairly large number of words from which one could potentially analogize to a productive pattern without any consequent productivity.'' (p. 25).
%
%
Thus, we need to complement the type count (henceforth $V$) with other measures to gauge how productive a particular inflectional class is. One such measure is the average (or median) token frequency of word types in a class.  The higher the average frequency of use is, the more likely it is to be less productive. Unproductive types typically have fewer lexemes, with lemma frequencies that tend to be high.  If their frequency were low, they would likely shift to a more productive type \citep[][]{lieberman2007quantifying}. In what follows, \texttt{Median} denotes the median frequency of the lemmas in a given inflectional class.

Two other measures for productivity build on the Good-Turing estimate for the probability of unseen types \citep{Good:1953}, estimated by the ratio of word types with frequency 1, the hapax legomena $V(1)$, and the total number of tokens $N$:
\begin{equation}
p = \frac{V(1)}{N}.
\end{equation}
The probability $p$ in (\theequation) can be understood as the rate at which the vocabulary size observed for a sample of $N$ tokens is increasing at sample time $N$. When considering the growth rate of the total vocabulary, the count of hapax legomona (henceforth $V(1)$) in a given class is proportional to the contribution of that class to the vocabulary growth rate . The ${\cal P}^\ast$ measure \citep{Baayen:92:YoM} is the ratio of $V(1)$ and the total number of hapax legomena calculated across all inflectional classes. Alternatively, one can focus on a given inflectional class, and estimate the probability that when another token of this class is sampled, it will be a token of a previously unseen type. This leads to the productivity measure $\cal P$ \citep{Baayen:91:YoM}, with $V(1)$ and $N$ in (\theequation) now both restricted to words from a given inflectional class. 

\begin{figure}
    \centering
    \includegraphics[width=0.7\linewidth]{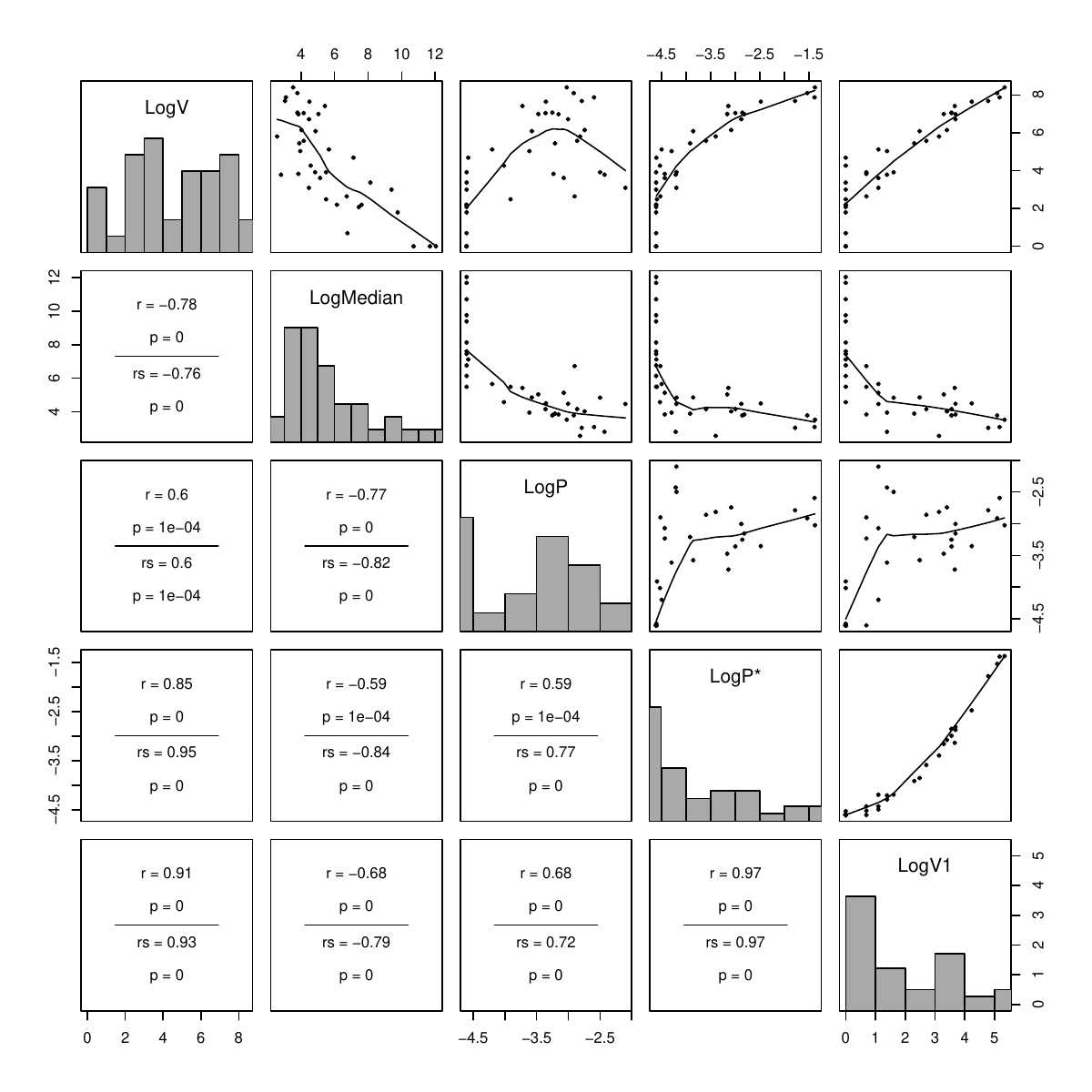}
    \caption{Pairs plot of 5 log-transformed measures gauging aspects of the productivity of Finnish inflectional classes.}
    \label{fig:pairscor}
\end{figure}

Figure~\ref{fig:pairscor} presents a pairwise scatterplot matrix for the 5 measures gauging productivity. All measures were log-transformed, backing off from zero by 1 for $V(1)$ and by 0.01 for $\cal P$ and ${\cal P}^\ast$.  This figure is based on the measures provided by \cite{nikolaev2008suomen} for all Finnish nominal inflectional classes for which hapax legomena were attested in Kielipankki (www.csc.fi) newspaper corpus, which sampled 131,406,087 words of written Finnish.  For eight inflectional classes, the number of types was so small that further measures are meaningless: inflectional class \#19 (exemplifying word \textit{suo}), \#31 (\textit{kaksi}), \#30 (\textit{veitsi}), \#29 (\textit{lapsi}), \#35 (\textit{lämmin}), \#37 (\textit{vasen}), \#42 (\textit{mies}), \#44 (\textit{kevät}), and \#46 (\textit{tuhat}).  These inflectional classes are not included in the dataset on which Figure~\ref{fig:pairscor} is based.   What Figure~\ref{fig:pairscor} clarifies is that all measures are positively correlated, with as exception \texttt{LogMedian}, which, as expected, enters into negative correlations with all other measures. The less productive an inflectional class is, the more the words of such a class depend on intense frequency of use to withstand the pressure of regularization.

\subsection{Cognitive status of inflectional classes}

In the 20th century, the return to inflection classes in Finnish linguistics was typically motivated by the idea that inflection classes are highly practical for teaching Finnish to second language learners \citep{raun1959suomen, schot1994taivutuksen}. Inflectional classes also help the analyst to document areas of systematicities.  Inflectional classes are also indispensable for developing finite state grammars \citep{roark2007computational} or DATR-based \citep{Evans:Gazdar:96} grammars. Unfortunately, the development of such grammars requires considerable hand-crafting, and somewhat arbitrary decisions about how to set up inflectional classes that make sense in that they are general enough but also do not have too many exceptional subclasses.  The history of Finnish linguistic investigations into nominal inflection illustrates the different kinds of systems that analysts can decide on. 

What then is the cognitive reality of the Finnish inflectional classes for native speakers?  Here, we leave aside what the consequences are of having received explicit instruction in the classroom about how to properly inflect Finnish nouns, and focus on the implicit knowledge that native speakers acquire over their lifetimes.   One possible approach to answering this question is to choose as point of departure some form of realizational morphology, or some finite state machine or DATR grammar, and argue that these models provide a high-level blueprint of the speaker's cognitive architecture.  The fact that we observe differences in the productivity of inflectional classes by itself can be argued to bear witness to the cognitive reality of these classes. In support of this approach, one could also point to a range of studies addressing the psychological reality of stems and exponents \citep{Niemi.etal:94, jarvikivi2002allomorphs, jarvikivi2002form, Jarvikivi:2003, hovdhaugen2000history}, as well as studies addressing the relevance of inflectional paradigms and the extent to which the different paradigm members of a lexeme are in actual use \citep{Loo:2018a, Loo:2018b}. \textcolor{black}{Furthermore, computational experiments by \cite{janda2021less} {\color{black} addressing form-to-form mappings for inflection} suggest that training on full paradigms (the basis for class definition) can be less effective for generalization than training on the more ecologically valid input of single, frequent word forms. This provides computational evidence that learners might achieve mastery despite, rather than because of, the existence of complete, abstract paradigmatic structures.}

In what follows, we take a different approach and investigate whether a cognitively motivated computational model of the mental lexicon, the Discriminative Lexicon Model \citep[DLM, ][]{Heitmeier:Chuang:Baayen:2024}, can learn to understand Finnish inflected nouns without any explicit information about stems and exponents, case and number, gradation, and inflectional classes. Specifically, we are interested in whether this model will understand novel forms from more productive inflectional classes better than novel forms from less productive or unproductive inflectional classes.  These questions are addressed in the next section.

%

\section{Error-driven subliminal learning of Finnish nominal inflection}

\subsection{Method}

The discriminative lexicon model (DLM) provides an error-driven computational framework for the modeling of lexical processing.  Word's forms are represented by high dimensional form vectors, the row vectors of a matrix $\bm{C}$, and words' meanings by high-dimensional meaning vectors (embeddings) brought together as the row vectors of a matrix $\bm{S}$.  In this study, we make use of fasttext embeddings \citep{Bojanowski:2017:fastText} to represent words' meanings. Thus, each row vector of $\bm{S}$ is a 300-dimensional fasttext embedding. To represent words' forms, we make use of multiple-hot 1/0 vectors that specify, for a given word, which of the $k$ n-grams or $n-phones$ attested across all the words in the data are present in that word.  For our dataset, there are 3103 different letter 3-grams, and 13,364 different letter 4-grams. In previous work with the DLM, 3-grams have been widely used \citep[see, e.g.,][]{Heitmeier:Chuang:Baayen:2024}, but, taking inspiration from \citet{snell2024pong}, in the present study, we also consider letter 4-grams.  

The simplest mappings of the DLM are linear mappings, a mapping $\bm{F}$ for comprehension, estimated from the equation
$$
\bm{C}\bm{F} = \bm{S}
$$
and a mapping $\bm{G}$ for production, estimated from
$$
\bm{S}\bm{G} = \bm{C}.
$$
Instead of using linear mappings, one can make use of multi-layer deep networks.  Both types of mappings are implemented in the \textbf{JudiLing} package \citep{Heitmeier:Chuang:Baayen:2024}.
 
Mappings can be trained type-wise or token-wise.  When using type-wise training, the rows of the $\bm{C}$ and $\bm{S}$ represent word types. Type-wise training examines what can be learned from a type list with infinite experience.  We will refer to linear mappings that are type-based as EOL (endstate-of-learning) mappings.  Alternatively, mappings can be based on matrices that have as many rows for a given type as there are tokens of this type in a corpus. Such matrices become huge and severely collinear.  Fortunately, the consequences of having multiple tokens per type for learning a linear mapping can be effectively handled by means of frequency weights \citep[see][for a proof]{Heitmeier:Chuang:Axen:Baayen:2023}.  We will refer to mappings that take token-frequency into account as mappings that use frequency-informed learning (FIL).

One way of assessing the quality of a mapping is to inspect how accurate the mapping is.  Does a mapping predict the right meaning for a given form, and the correct form given a meaning?  We evaluate correctness by examining whether the nearest neighbor of a predicted vector is the `gold standard' vector that has to be learned.  For instance, if a comprehension mapping takes the form vector \texttt{dog} and places it very close to the semantic vector \textsc{dog} and far away from \textsc{cat}, the mapping of good quality.  But if \texttt{dog} places its predicted semantic vector much closer to \textsc{cat} than to \textsc{dog}, the mapping is of low quality.   

It is often insightful to evaluate how accurate a mapping is for the words it has been trained on.  However, as a mapping can be overfitting the data, it is also useful to ascertain how well a mapping performs under cross-validation. The mapping is trained on, for instance, 90\% of the data, and evaluated on the remaining 10\% of the data that it has not seen during training.  In this way,  we can assess how well the model generalizes to unseen data. In other words, with cross validation, we are assessing whether a given mapping is productive.  How productive the mapping is will depend on the extent to which there are regular and systematic correspondences between the form space and the meaning space.

\subsection{Materials}

In this study, we draw on the same dataset described in \citet{nikolaev2022generating}, which contains 2,000 high-frequency Finnish nouns (excluding compounds) sourced from a frequency lexicon of Finnish newspaper text (\url{http://urn.fi/urn:nbn:fi:lb-201405272}). All available inflected forms of these nouns were extracted from a Finnish corpus (84,308,641 tokens) of online discussions in a Reddit-like community (\url{http://urn.fi/urn:nbn:fi:lb-2017021505}), yielding 104,716 distinct forms (10,427,959 tokens, 12.4\% of the corpus). Syncretic forms (10.3\%) were disambiguated by choosing the function with the highest corpus frequency (see \citealp{nikolaev2022generating} for details). We then obtained pre-trained {\tt fasttext} embeddings \citep{grave2018learning} from \url{https://fasttext.cc} for 55,271 of these 104,716 inflected forms.

\subsubsection{Distributional semantics of Finnish inflectional classes}\label{sec:LDA}

Before examining whether error-driven learning is sensitive to the productivity of inflectional classes, we first consider the embedding space in some more detail. What lexical properties are represented in this embedding space? Previous research on Finnish nominal inflection has shown that case can be predicted with high accuracy from fasttext word embeddings, and that the semantics of plurality vary within case \citep{nikolaev2022generating}.  Given that inflectional class has been found to also be predictable to a considerable extent, as shown by \citet{Orzechowska:Baayen:2025} for Polish \citep[see also][]{williams2019quantifying}, we consider whether Finnish embeddings also predict inflectional class. If this is indeed the case, then there must be consistent variation in the semantic space that correlates with the form properties that characterize the inflectional classes.  A positive answer would also contribute to improving our understanding of the performance of error-driven learning with the DLM model. 

We made use of Linear Discriminant Analysis (LDA) to predict inflectional class from fasttext embeddings. The LDA was run on a dataset from which 312 words without an identifiable inflectional class were excluded. Accuracy was at 48\%. Under leave-one-out cross validation, accuracy dropped by only 2\% to 46\%.  Both accuracies are way above the majority baseline of 16\%. This indicates that the embeddings contain considerable information about inflectional classes. 

It should be noted that even though LDA can predict inflectional class from fasttext embeddings surprisingly well, these embeddings are far more informative about other morphological features. As shown in \citet{nikolaev2022generating}, case and number information can both be predicted using LDA with accuracies around 96\%, and for the combined category of case and number, accuracy is very similar, at 94\%. Clearly, information about case and number is much better represented in embeddings than information about inflectional class. 

To further understand how embeddings encode class information, in addition to case and number, we calculated the plural shift vectors for the singular-plural noun pairs in our dataset. These shift vectors are defined as the plural embedding from which the corresponding singular embedding is subtracted, and hence specify how in semantic space to move from a singular to its corresponding plural.  For our data, we found 13,285 word pairs which have the same lexeme and case but differ in number.  LDA can predict inflectional class from these shift vectors with an accuracy of 53\%, which is slightly better than the results obtained using plain embeddings instead of shift vectors. However, under leave-one-out cross validation, the accuracy for shift vectors, 44\%,  is similar to that of raw embeddings. Furthermore, we found that case information is completely preserved in number shift vectors: an LDA classifier is able to predict case perfectly (99.6\%), even for unseen token (98.5\% under leave-one-out cross validation).

Given that case and number are so well represented in the embeddings, we considered the possibility that when case and number are controlled for, inflectional class can be predicted with higher accuracy. To address this possibility, we ran an LDA separately for each case and number combination (e.g., nominative singular, allative plural). Results are presented in Table \ref{tab:lda}. As can be seen, there is a substantial increase in accuracy when inflectional class is predicted within a given number-case category, with accuracies ranging from 64.1\% to 100\% without cross-validation, and from 40.3\% to 67.2\% with leave-one-out cross-validation.  These findings lead to the conclusion that the semantics associated with inflectional class vary with case and number, and thus are to a considerable extent specific to individual paradigm cells.  These findings also highlight that mappings between form and meaning face a highly complex task of linking highly variable form properties to semantic vectors in which case, number, and inflectional class interact.

\begin{table}[]
    \centering
    \begin{tabular}{l|rrrrrr} \hline
        number\_case&n class&n lexeme&n token&LDA(\%)&LDA  CV(\%)&majority baseline(\%) \\ \hline
        sg\_abe&16&105&123&NA&NA&NA \\
        sg\_abl&36&1192&1566&91.8&65.8 &15.5 \\
        sg\_ade&38&1761&2735&82.8&63.1&16.2 \\
        sg\_all&37&1622&2288&87.6&66.1&15.1 \\
        sg\_ela&38&1897&3249&81.4&63.2&16.3 \\
        sg\_ess&36&1429&1747&89.2&61.5&14.7 \\
        sg\_gen&38&1973&3260&70.3&46.9&15.1 \\
        sg\_ill&38&1854&3001&80.3&61.0&17.1 \\
        sg\_ine&38&1609&2852&82.3&62.5&16.8 \\
        sg\_nom&38&1979&5223&64.1&49.2&17.6 \\
        sg\_par&38&1948&4153&76.2&60.5&17.7 \\
        sg\_tra&36&1433&1831&91.5&67.2&16.2 \\
        sg\_pl\_com&30&665&685&99.3&60.1&16.2 \\
        pl\_abe&14&61&76&NA&NA&NA \\
        pl\_abl&30&689&816&98.0&65.9&19.1  \\
        pl\_ade&36&1308&1729&90.7&63.2&16.1  \\
        pl\_all&33&1184&1558&91.1&62.3&16.8 \\
        pl\_ela&35&1611&2398&83.4&59.8&15.7 \\
        pl\_ess&30&768&823&97.8&60.0&17.5 \\
        pl\_gen&36&1678&3108&78.3&58.0&15.4 \\
        pl\_ill&34&1486&2023&89.0&64.4&14.3 \\
        pl\_ine&36&1297&1872&87.5&61.5&15.5 \\
        pl\_ins&30&463&496&100.0&40.3&17.3 \\
        pl\_nom&38&1867&3323&69.6&47.3&15.7 \\
        pl\_par&36&1757&3317&76.5&57.3&15.0 \\
        pl\_tra&26&650&707&99.3&62.2&19.1 \\ \hline
    \end{tabular}
    \caption{LDA accuracies, LDA accuracies under leave-one-out cross validation, and majority baseline accuracies for each case-number combination. Results are not reported for categories with very few available tokens.}
    \label{tab:lda}
\end{table}

In what follows, we first discuss computational experiments addressing comprehension, which, as will become clear, is the easier task. We then turn to the modeling of production, the more difficult task.

\subsection{Comprehension experiments}

We implemented four DLM models, two trained with EOL and two with FIL. For each learning type, we fitted a model to the data using letter 3-grams, and a model using letter 4-grams.  In what follows, we first report overall accuracies.  We then investigate accuracy by inflectional class. We conclude with an analysis of accuracy at the word level.

\subsubsection{Accuracy on training and test data}

Table~\ref{tab:comp_acc} summarizes the performance of the four models for training as well as held-out test data.  The training data comprised the 40,694 words with a frequency greater than 5.  The test data comprised the 14,577 words with a frequency less than or equal to 5. We first consider accuracy for models based on form vectors with 3-grams. 

The percentage of types that were correctly recognized (type accuracy @1) is higher for training data than for test data, for both EOL and FIL.  As the present classification task, with 55,271 different candidates, is far from trivial, we also calculated the proportion of cases where the target is among the top 10 candidates (accuracy @10). Unsurprisingly, these accuracies are higher than the accuracies@1.  Evaluated in this way, FIL accuracy is more similar to EOL accuracy, both for training data and for test data.  In other words, FIL gets close, but lacks precision for type-based evaluation.

Since usage-based models trained with FIL learn higher-frequency words better than lower-frequency words, it is also informative to weight accuracies for the token frequencies of the types.  Token-based accuracy on the training data with FIL is at 84\%, which is substantially higher than token-based accuracy for EOL (73\%).  With EOL, more lower-frequency words are learned precisely, but this happens at the cost of accuracy for some higher-frequency words. 

Next, consider accuracy for the models based on form representations built on 4-gram. Interestingly, these models  substantially outperform models with letter 3-grams not only on the training data, but also on the test data.  Accuracy@1 on the held-out data for FIL nearly doubles, increasing from 36.15\% to 64.29\%.  Given these results, we expect that the models using 4-gram vectors will also align better with the different degrees of productivity of the Finnish inflectional classes.


\begin{table}[htbp]
    \centering
    \begin{tabular}{llrrrrr} \hline 
         &   &   \multicolumn{3}{c}{training data} & \multicolumn{2}{c}{test data}      \\ 
         &   &  \multicolumn{2}{c}{type}            &  token     &  \multicolumn{2}{c}{type}             \\ \hline
         & grams  &  @1       & @10        &            &  @1    &  @10  \\ \hline
     EOL &     3  &  82.46    & 95.36      &  72.87     &  78.05 & 96.83 \\ 
     EOL &     4  &  98.40    & 99.99      &  97.84     &  85.95 & 99.65 \\ \hline
     FIL &     3  &  49.93    & 84.22      &  84.22     &  36.15 & 83.49 \\
     FIL &     4  &  75.96    & 98.37      &  95.76     &  64.29 & 98.29   \\ \hline
    \end{tabular}
    \caption{Comprehension accuracy for EOL and FIL evaluated on training and test data, for models with 3-gram and 4-gram form vectors. Type: accuracy calculated for types; token: accuracy calculated for tokens. @1: target is the best candidate; @10: target is among the top 10 candidates. Held-out words had a frequency less than or equal to 5, and constitute 26.4\% of the dataset.}
    \label{tab:comp_acc}
\end{table}

For our dataset, the Good-Turing estimate of encountering a token of a novel, previously unseen type is extremely small: 0.0005. This observation dovetails well with \citet{karlsson1986frequency} corpus-based study, which reports that  though in theory a word can appear in any of the inflected forms defined by grammar, only a subset of the possible forms are actually in use \citep[see also ][]{janda2021less}. Thus, from a cognitive perspective, accuracy on training data is likely to be more realistic than evaluation on held-out data, at least for adult language users.


\subsubsection{Accuracy for held-out data}

\begin{sidewaysfigure}
    \centering
    \includegraphics[width=1.0\linewidth]{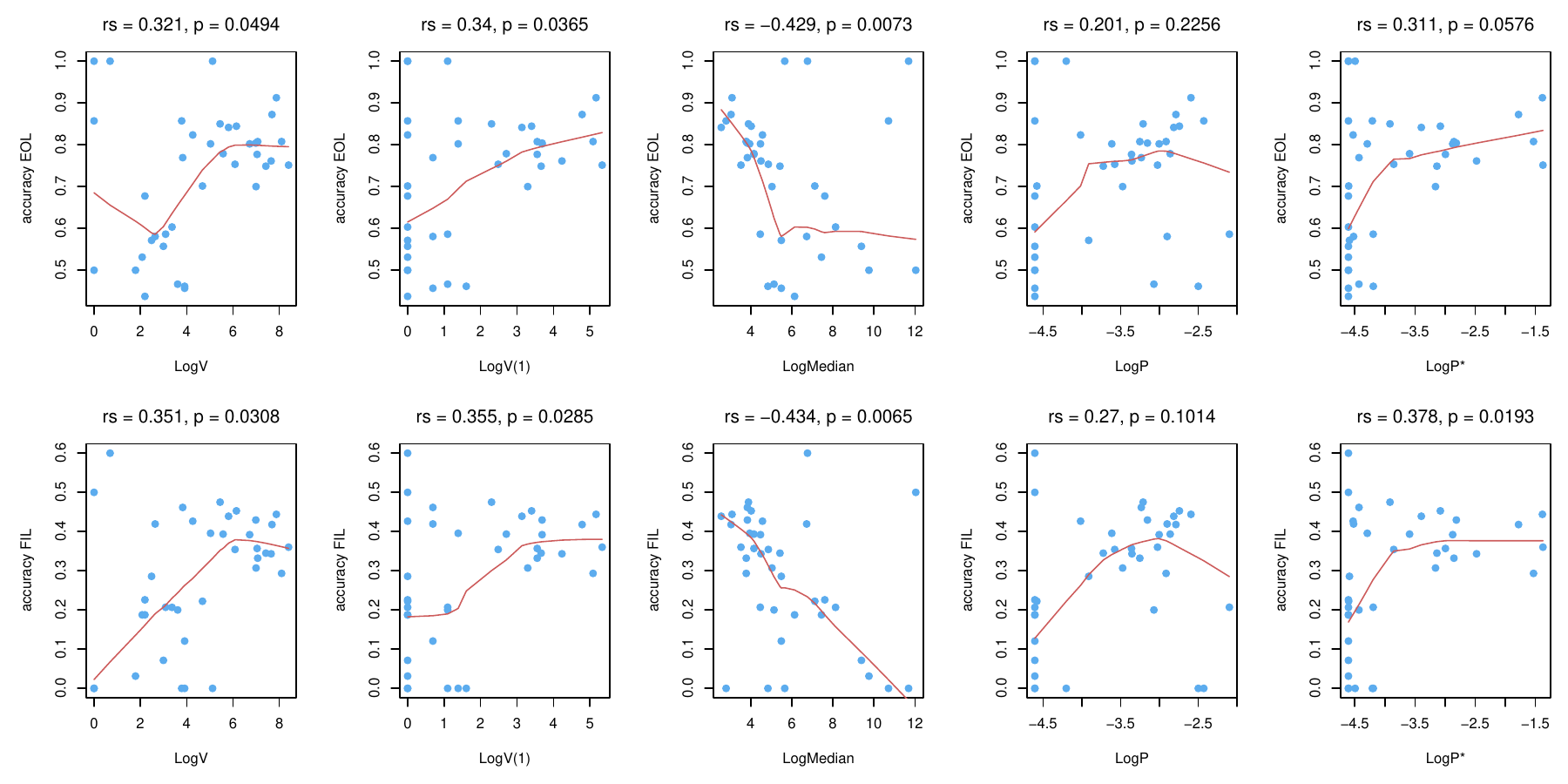}
    \caption{EOL (top) and FIL (bottom) comprehension accuracies plotted against productivity measures, for models using 3-grams to represent words' forms. Data points represent inflectional classes.}
    \label{fig:comprehension_scatterplots}
\end{sidewaysfigure}

\begin{sidewaysfigure}
    \centering
    \includegraphics[width=1.0\linewidth]{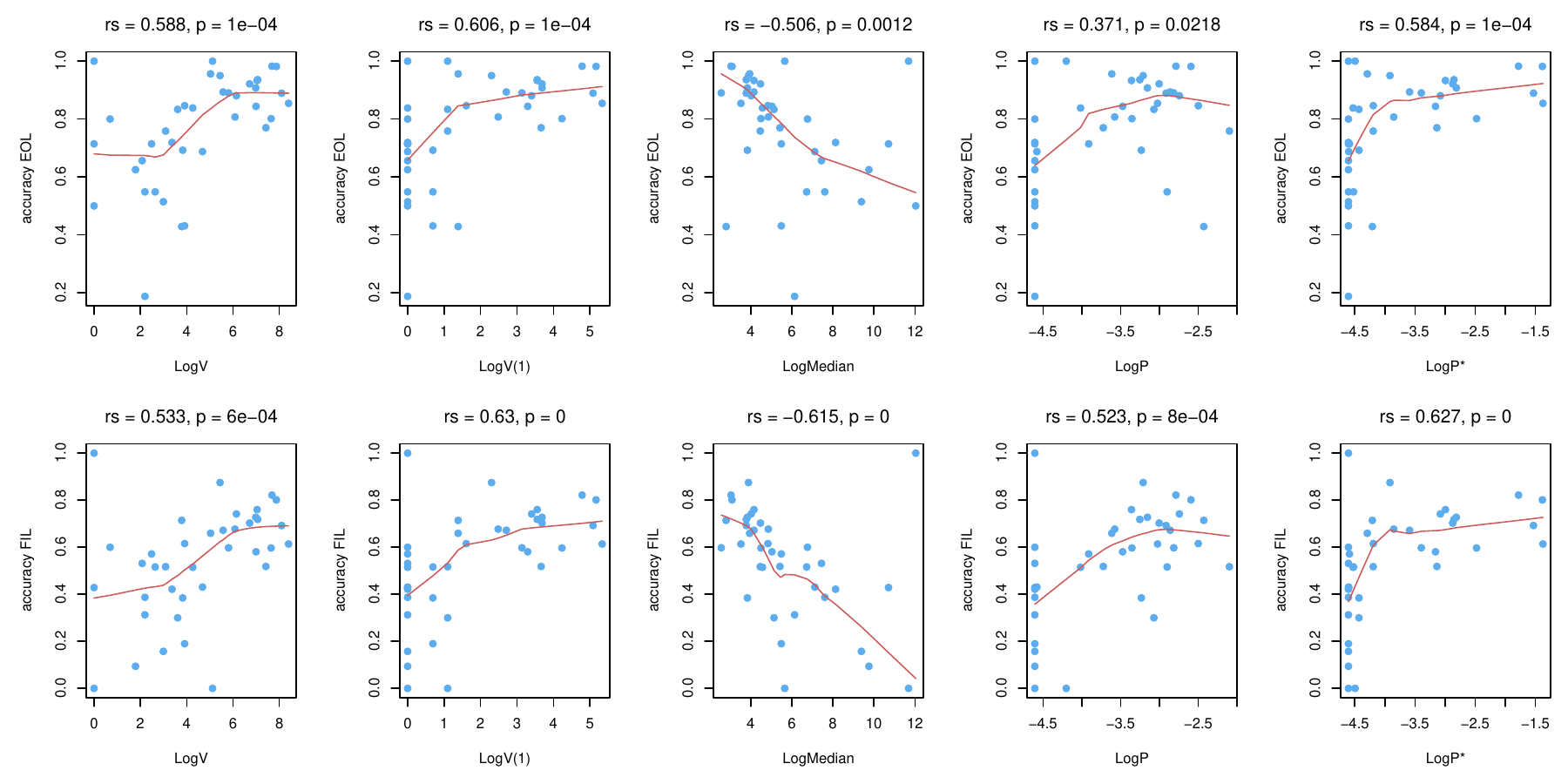}
    \caption{EOL (top) and FIL (bottom) comprehension accuracies plotted against productivity measures, for models using 4-grams to represent words' forms. Data points represent inflectional classes.}
    \label{fig:comprehension_scatterplots4}
\end{sidewaysfigure}

Figure~\ref{fig:comprehension_scatterplots} presents, for the models using 3-gram form vectors, scatterplots for the correlations of type-based accuracy@1, for the held-out data, for EOL (top panels) and FIL (bottom panels) with the 5 productivity measures introduced above.  All trends are noisy, but Spearman correlation tests for EOL provide tentative support for the expected positive trends for \texttt{LogV}, \texttt{LogV(1)}, and the expected negative trend for \texttt{LogMedian}. In the case of FIL, some support is also present for \texttt{LogP*}. Overall, patterns appear to be slightly less noisy when the model is trained with FIL.  Figure~\ref{fig:comprehension_scatterplots4} presents the corresponding scatterplots for the model trained with form vectors based on 4-grams. All correlations have improved substantially.  Absolute correlations for FIL again tend to be higher than those for EOL, with as only exception the correlation for \texttt{LogV}.

Considered jointly, these results indicate that a discriminative error-driven learning model for comprehension can learn the words of the more productive inflectional classes with greater precision.  In other words, we now have evidence that productivity and generalization in error-driven learning go hand in hand. 
These results align well with the findings of 
\citealt{janda2021less} a model trained only on frequency-ordered single forms eventually surpassed a full-paradigm model in accuracy on unseen forms, suggesting that focusing learning resources on high-probability occurrences is an effective strategy for generalization.


\subsubsection{Statistical evaluation with control variables}

In section~\ref{sec:LDA}, we observed that the extent to which inflectional class is predictable varies by case and number. It is therefore conceivable that the results obtained in the previous section are confounded with case, number, and their interaction.  A further possible confounding factor is the frequencies of the lexemes.  We therefore fitted two generalized additive models \citep{Wood:2017}, one to words' target correlations and one
to words' accuracies.  The target correlations are the model's estimate of the correlation of the predicted semantic vector and the targeted gold standard fasttext embedding, its target correlation.  A word is accepted as accurately understood when its target correlation is higher than its correlation with any other fasttext embedding. As predictors, we included log lexeme frequency, case, number, and the interaction of case and number, and inflectional class.  Models that included the interaction of case and number provided superior fits to the data (117 AIC units for the Gaussian GAM for the target correlations, and 22 AIC units for the logistic GAM for the accuracies). We modeled the case-number combinations, and inflectional class as random-effect factors.

\begin{figure}
    \centering
    \includegraphics[width=0.8\linewidth]{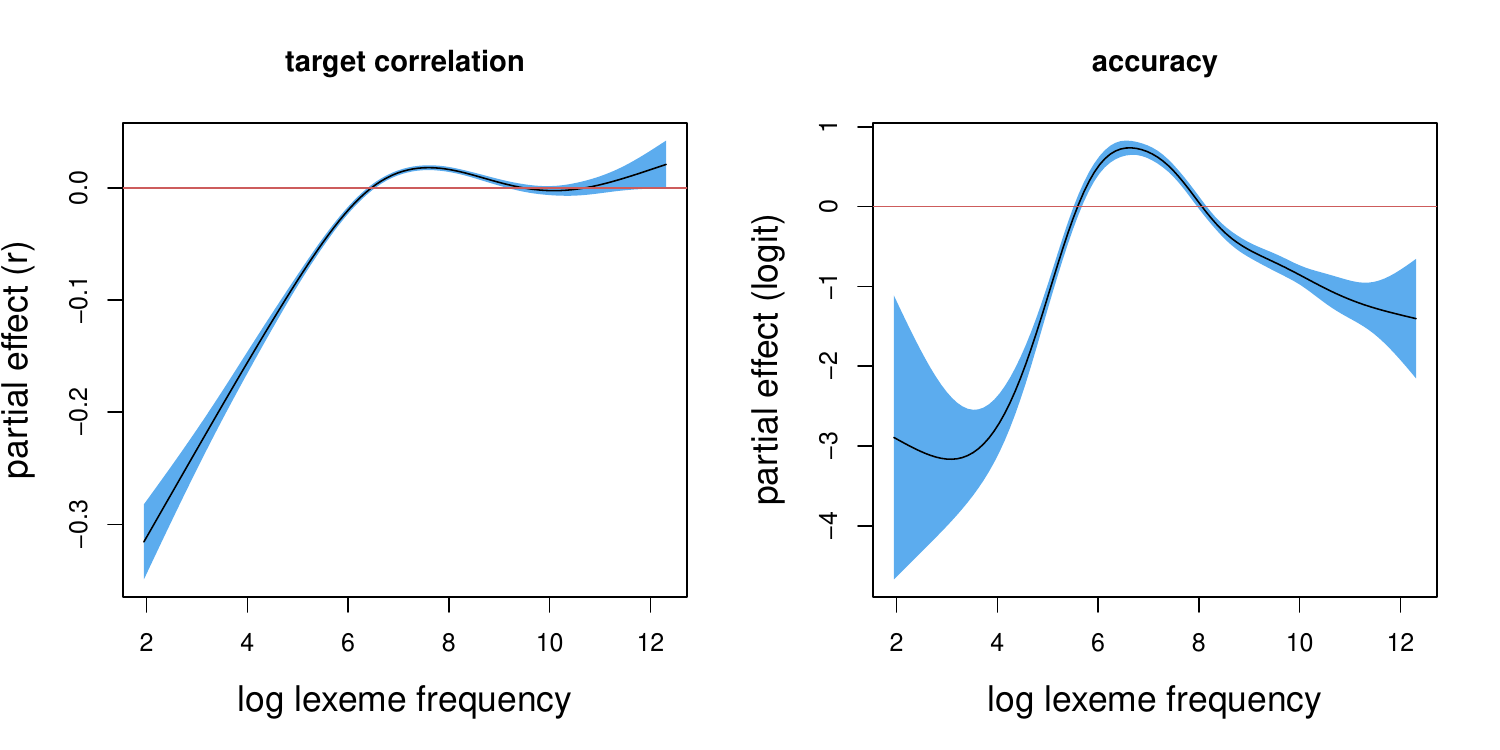}
    \caption{Partial effect of log lexeme frequency in a GAM fitted to the target correlations (left panel) and a logistic GAM fitted to the accuracies (right panel), for a \textbf{usage-based comprehension} model trained with FIL.}
    \label{fig:comprFIL4corLogit}
\end{figure}

Figure~\ref{fig:comprFIL4corLogit} presents the partial effect of log lexeme frequency in the two GAM models.  Inflectional variants of lower-frequency lexemes are not predicted with precision.  For the target correlation, the frequency effect levels off halfway through the range of frequency values, whereas for the accuracies, its effect reverses, possibly because higher-frequency lexemes are more likely to have formal or semantic irregularities.


\begin{table}[htbp]
\centering
\caption{Spearman correlations of by-class random intercepts and 5 productivity measures, for a Gaussian GAM fitted to the target correlations of the \textbf{FIL-based comprehension} model, and for a logistic GAM fitted to the corresponding accuracy data.}
\label{tab:compSpearmans}
\begin{tabular}{lrrrr} \hline
           &   \multicolumn{2}{c}{target correlation} &   
               \multicolumn{2}{c}{accuracy}             \\ \hline
measure    & estimates & p-value  &   estimates & p-value  \\ \hline
LogP       &  0.35     & 0.0307   &   0.46      & 0.0040    \\
LogV       &  0.36     & 0.0278   &   0.51      & 0.0010    \\
LogV1      &  0.43     & 0.0070   &   0.61      & $<$ 0.0001\\
LogMedian  & -0.40     & 0.0141   &  -0.54      & 0.0004    \\
LogPstar   &  0.44     & 0.0061   &   0.60      & $<$ 0.0001\\ \hline
\end{tabular}
\end{table}

We also extracted the by-class random intercepts and correlated these with the five productivity measures, using the Spearman correlation.  Table~\ref{tab:compSpearmans} provides a summary of the results. Estimated correlations are higher for the accuracy GAM, compared to the target correlation GAM.  As these results are similar to those reported in Figure~\ref{fig:comprehension_scatterplots4}, we conclude that the learnability of inflectional classes is not confounded with lexeme frequency nor with case and number.

\subsection{Production experiments}

\noindent
In the standard set-up of a discriminative lexicon model for production, the first step in the production process is to predict what n-grams are supported by a word's meaning (an approximation of the `what' system). As a second step, the n-grams are ordered into words (an approximation of the `where' system).  The `where'-algorithm produces a set of candidates for production, and selects that candidate form that, when given as input to the comprehension mapping, best approximates the meaning intended by the speaker. This internal feedback loop is referred to in the DLM model as synthesis by analysis.

For the first step in the production process, we need a mapping from 300-dimensional fasttext vectors to 3103-dimensional 3-gram based form vectors.\footnote{For production, 3-grams represent context-sensitive articulatory targets, resembling a common approach to production in articulatory phonology \citep{Browman:Goldstein:92}. Whereas 4-grams are motivated by research on reading \citep{snell2024pong}, 4-grams have no such motivation for production.  Furthermore, computing accuracies for form vectors built from 4-grams requires prohibitive amounts of memory.
}  
A linear mapping from a low-dimensional space into a high-dimensional space cannot be done with precision. A non-linear mapping with a multi-layer network is much better positioned for this task.  Non-linear mappings that do not take frequency of use into account, and hence approximate an endstate of learning similar to linear EOL mappings, are relatively straightforward to evaluate.  In  Section~\ref{sec:deepEOL}, we report the results of a deep discriminative learning (DDL) production model. It is possible in principle to take frequency of use into consideration, but this is not feasible for the present dataset. Such a model is computationally extremely demanding, as it would have to be trained step by step on some 10 million tokens.  One workaround is to thin the data by training not with the observed frequencies of use, but with a fraction of these frequencies, rounded to the nearest higher integer.  For the present study, this is not an option, as many lower frequencies would be rounded to 1, which is undesirable for a study of productivity.  Furthermore, words presented only once in a frequency-sensitive training set-up are extremely difficult, if not impossible, to learn. In Section~\ref{sec:LDLwith}, we therefore use FIL mappings to explore what the bottlenecks are for learning to produce Finnish nouns.

\subsubsection{Modeling without taking frequency into account: endstate learning}\label{sec:deepEOL}

For the `what' system, we made use of a 300 $\times$ 1000 $\times$ 3103 deep network (with in all 3,407,103 parameters) with rectified linear units for the hidden layers, and trained with the binary cross-entropy loss \citep[see][for an introduction to using these kind of networks within the DLM framework]{Heitmeier:Chuang:Baayen:2024}.  Trained for 100 epochs with early stopping resulted in a matrix with predicted form vectors $\hat{\bm{C}}$ that was very similar to the $\bm{C}$ matrix itself. Accuracy was at 100\%.
%
%
%

In order to assess how well this type of network generalizes, we split the dataset into a training dataset with 49,771 words, and a test dataset with 5500 words (about 10\% of the full dataset). The words in the test set were selected such that their lexeme, their case, and their number were also present in the training data. We fitted a deep network using the same parameter settings as used in the previous model, and investigated its accuracy for the training data (99.9\% correct) and the test data (87.2\%).   
For understanding the challenges posed to the model by the test data, we removed words with uncertain inflectional class, as well as words with ambiguous number. The resulting dataset that we submitted to further statistical scrutiny comprised 5385 words. For each word, we extracted from the model its correlation $r$ with the gold standard form vector, and set its accuracy to 1 if its correlation $r$ was larger than its correlation with any of the other gold standard form vectors.

\begin{figure}[p]
    \centering
    \includegraphics[width=1.0\linewidth]{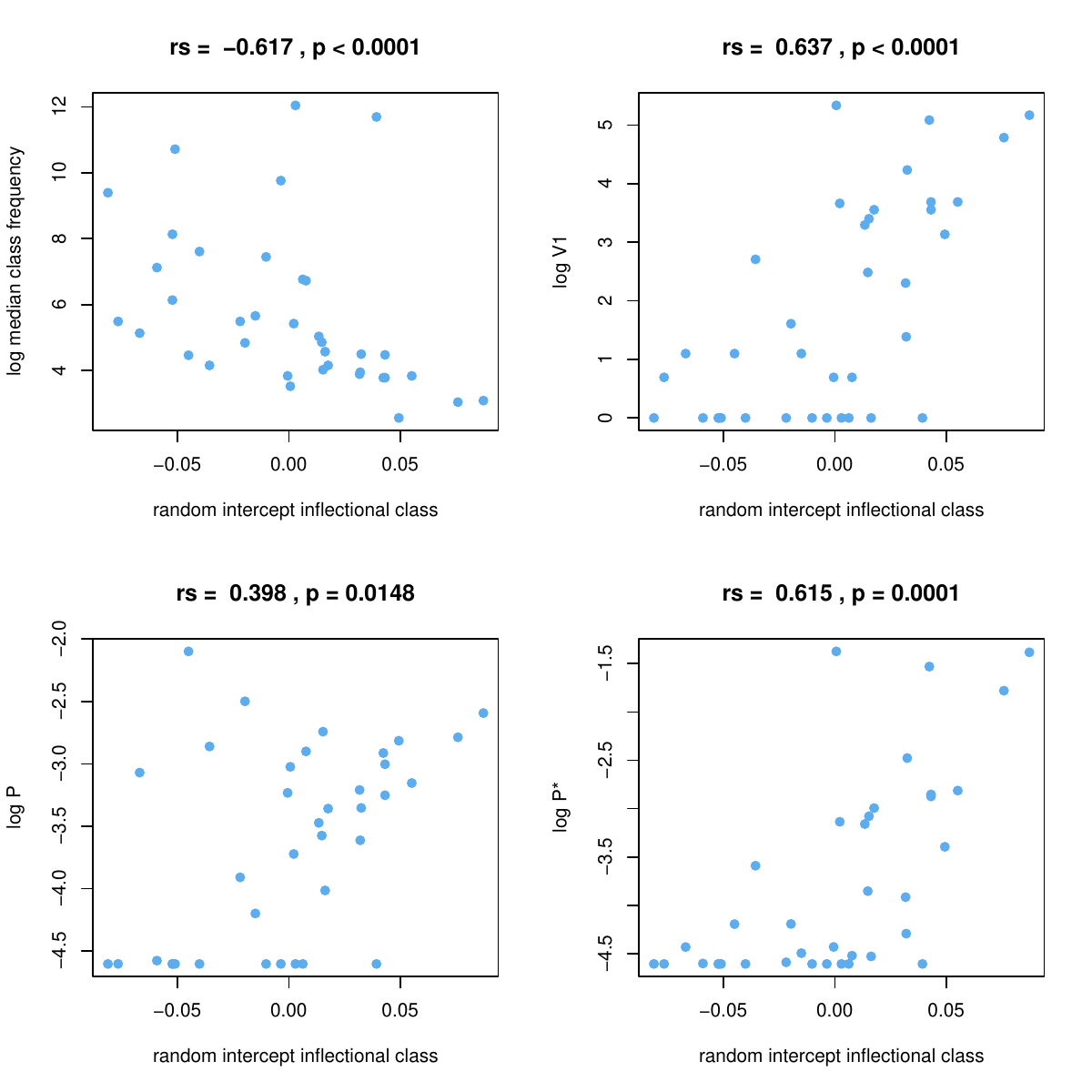}
    \caption{Scatterplots with Spearman correlations for the inflectional class random intercepts and four measures of inflectional class productivity, for the \textbf{`what'-system} modeled with \textbf{endstate learning}.  Random intercepts are taken from a QGAM fitted to the target correlations, i.e., the correlations between predicted and gold standard form vectors.}
    \label{fig:deepCeval}
\end{figure}

First consider the correlations $r$. We fitted a quantile GAM \citep{Fasiolo:Goude:Nedellec:Wood:2017} to these correlations, with random intercepts for Case and Inflectional Class, and a treatment contrast for Number.  As prediction of inflectional class using LDA improved substantially when carried out by paradigm cell (see Section~\ref{sec:LDA}), we also considered a model with random intercepts for all combinations of case and number.  The AIC of this second model was lower by 42.5 units, indicating that model fit improves when case and number are allowed to interact. From the QGAM, we extracted the random intercepts and used these as a measure gauging how well the words of an inflectional class can be learned, while taking the effects of case and number on learning into account.  Figure~\ref{fig:deepCeval} presents scatterplots for the random intercepts of the inflectional classes and four measures of class productivity.   Inflectional classes with high median class frequency are more difficult to learn, whereas more hapax legomena, a higher ${\cal P}$, and a higher ${\cal P}^{\ast}$ all afford higher learning accuracies.  

In summary, a deep mapping from embeddings to word forms achieves extremely high accuracy on training data, very good accuracy on test data (87.2\%).  Furthermore, accuracy varies by inflectional class ($p < 0.0001$) in a way that aligns with several productivity measures. 

The second step in the production process weaves the best supported trigrams (we used a support threshold of 0.01) into a set of word candidates, and selects that candidate for articulation that best realizes the intended meaning (the synthesis by analysis feedback loop).  We gave the algorithm as input the deep mapping from meaning to form, and a linear mapping from form to meaning, obtained with EOL applied to the training data.   We chose a linear mapping because its comprehension accuracy was satisfactory--98.4\% on the training data and 89.95\% on the test data. Production accuracy of the path weaving algorithm for the held-out data was 0.629.   

\begin{figure}[htbp]
    \centering
    \includegraphics[width=1.0\linewidth]{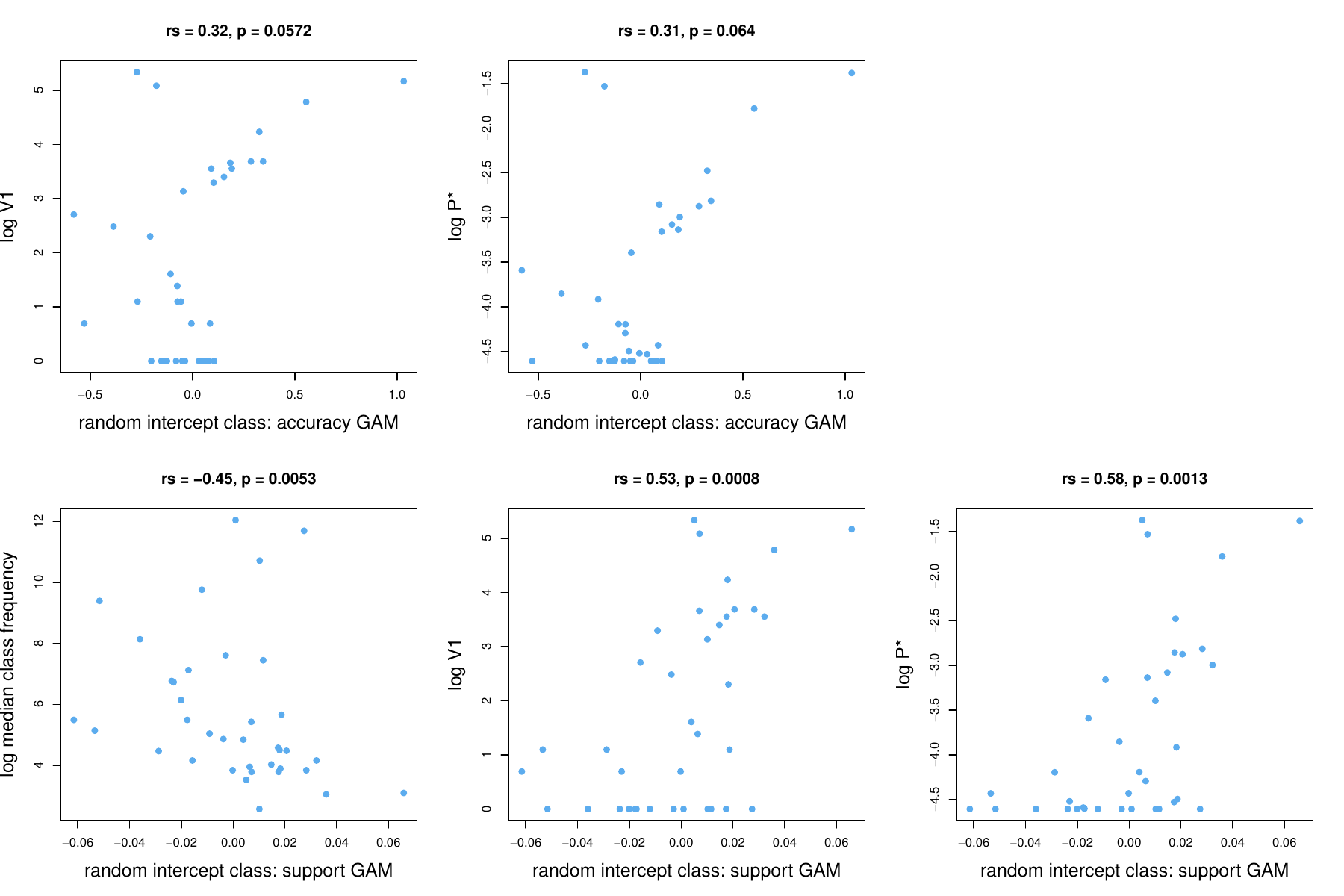}
    \caption{Scatterplots for the random intercepts for inflectional class and productivity measures, for the \textbf{where-system} modeled with \textbf{endstate learning}. For the upper panels, random intercepts are taken from a binomial GAM predicting the accuracy of the where system. For the lower panels, random intercepts for inflectional class are taken from a GAM predicting the support in the where system for the targeted word form.}
    \label{fig:deepWeave}
\end{figure}


Figure~\ref{fig:deepWeave}\ presents scatterplots for the random intercepts for inflectional class and productivity measures.  The upper panels present scatterplots for random intercepts in a binomial GAM predicting accuracy. They show a weak correlation with log V(1) and log ${\cal P}^{\ast}$. Instead of considering accuracy, we can investigate the support for a word's form, {\color{black}defined as the correlation between the semantic vector of the best candidate predicted by the feedback loop and the gold standard semantic vector. For example, if the target form is \textit{mato} `worm' and the predicted form is \textit{matto} `carpet', the support is the correlation of the predicted semantic vector of \textit{matto} and the gold standard semantic vector of \textit{mato}. In other words, the support measure gauges how accurately the form predicted the model matches the intended meaning, and is} a more fine-grained measure than the binary accuracy measure.  More convincing
correlations are visible in the lower panels of Figure~\ref{fig:deepWeave}.  As expected, the correlation is negative for the median class frequency, and positive for the other two measures.

\subsubsection{Modeling with FIL: usage-based  learning}\label{sec:LDLwith}

\noindent
The production model presented in the preceding section is a usage-free model.  This model is optimal for assessing what can be learned with infinite resources.  In what follows, we consider the consequences of taking frequency of use into account, using FIL linear mappings.  These mappings are much less accurate than the deep mapping, but more in line with usage-based theories, and promise to be informative about  potential learning bottlenecks in a cognitively more informed approach to lexical processing.
 
For the `what' system, we implemented a FIL linear mapping from embeddings to 3-gram form representations to get a sense of where simple linear mappings, as a proxy for human learning, face the greatest challenges.   For the `where' system, we started out with the empirical form vectors (equivalent to the form vectors predicted by the deep mapping) and used these as input to the `where'-algorithm that weaves n-grams into words.  For the synthesis by analysis feedback loop, a comprehension mapping trained with FIL was employed.  For computational convenience we used 3-grams to represent word forms (combining 3-grams for comprehension and 4-grams for production is currently not implemented in the JudiLing package).

For this experiment, we did not split the data in training and test data, as the low-frequency words provide a natural testbed for probing generalization to the less familiar forms.  As previous experience with FIL indicates that words with a frequency of 1 in training are not learnable, there is no reason to expect any better performance with held-out, zero-frequency words. 

As expected, a linear mapping from low-dimensional embeddings to high-dimensional form vectors has poor accuracy:  7\% accuracy@1 and 19\% accuracy@10.\footnote{A Linear mapping using EOL is also challenged, 10\% for accuracy@1, 33\% for accuracy@10.}  This is unsurprising, given that the network is given the task to map points in a lower-dimensional space onto points in a much higher-dimensional space.  However, the linear mappings are nevertheless useful for assessing learning difficulties. 

For each word, we calculated the target correlation, i.e., the correlation of that word's predicted form vector $\hat{\bm{c}}$ with its gold standard target form vector $\bm{c}$.  We also extracted from the model whether the predicted form was correct, i.e., that  the target correlation is at least as high as the correlation with the form vector of any other word.  We then fitted a Gaussian Location Scale model to the target correlations, allowing the variance to vary nonlinearly with word frequency, and predicting the mean from word frequency, the case by number combinations, and inflectional class.  All predictors were well-supported (all $p < 0.0001$).  A model including an interaction of Case by Number outperformed a model with only main effects for Case and Number by 2341 AIC units. 

\begin{figure}
    \centering
    \includegraphics[width=1.0\linewidth]{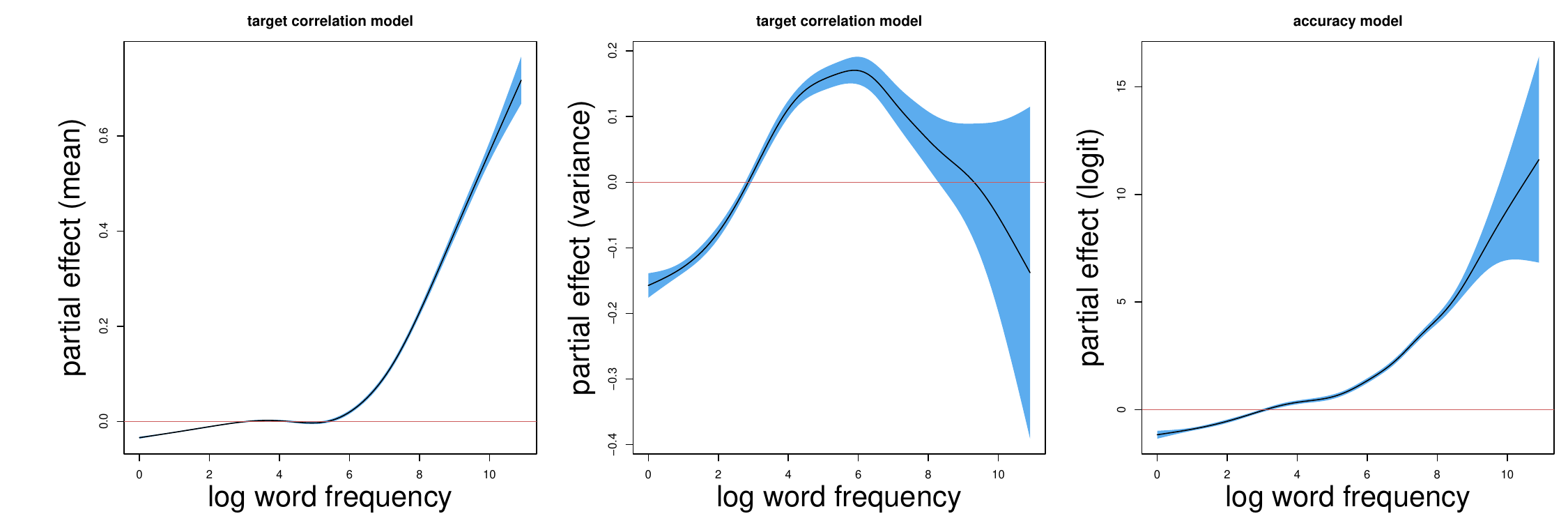}
    \caption{Partial effect of word frequency in the \textbf{what-system}, using \textbf{usage-based learning with FIL}. The left and center panels represent thin plate regression smooths in a Gaussian Location Scale GAM fitted to the target correlations. The right panel presents the partial effect of frequency in a binomial GAM.}
    \label{fig:prodFILwhat}
\end{figure}

The partial effect of frequency on mean and variance is presented in the first two panels of Figure~\ref{fig:prodFILwhat}. As expected, high-frequency words are learned well, in contrast to low-frequency words. A logistic GAM fitted to the accuracy data likewise calls attention to the importance of word frequency. Interestingly, the random intercepts for inflectional class are not correlated with any of the productivity measures.  Once frequency of use is taken into account, it dominates model performance, together with case and number. \textcolor{black}{\cite{janda2021less}'s results offer a potential explanation of why frequency dominates performance in FIL production models, weakening productivity correlations: if the learning system optimizes based on frequent, partial input (as their single-form model does), then frequency itself becomes the primary driver, and 
productivity measures calculated for abstract inflectional classes 
become less predictive for DLM  usage-based production models.}  The FIL meaning-to-form mapping of the what-system emerges as having very different properties than the FIL form-to-meaning mapping in comprehension, for which productivity measures were correlated with by-class accuracies.

\clearpage


\begin{figure}
    \centering
    \includegraphics[width=0.9\linewidth]{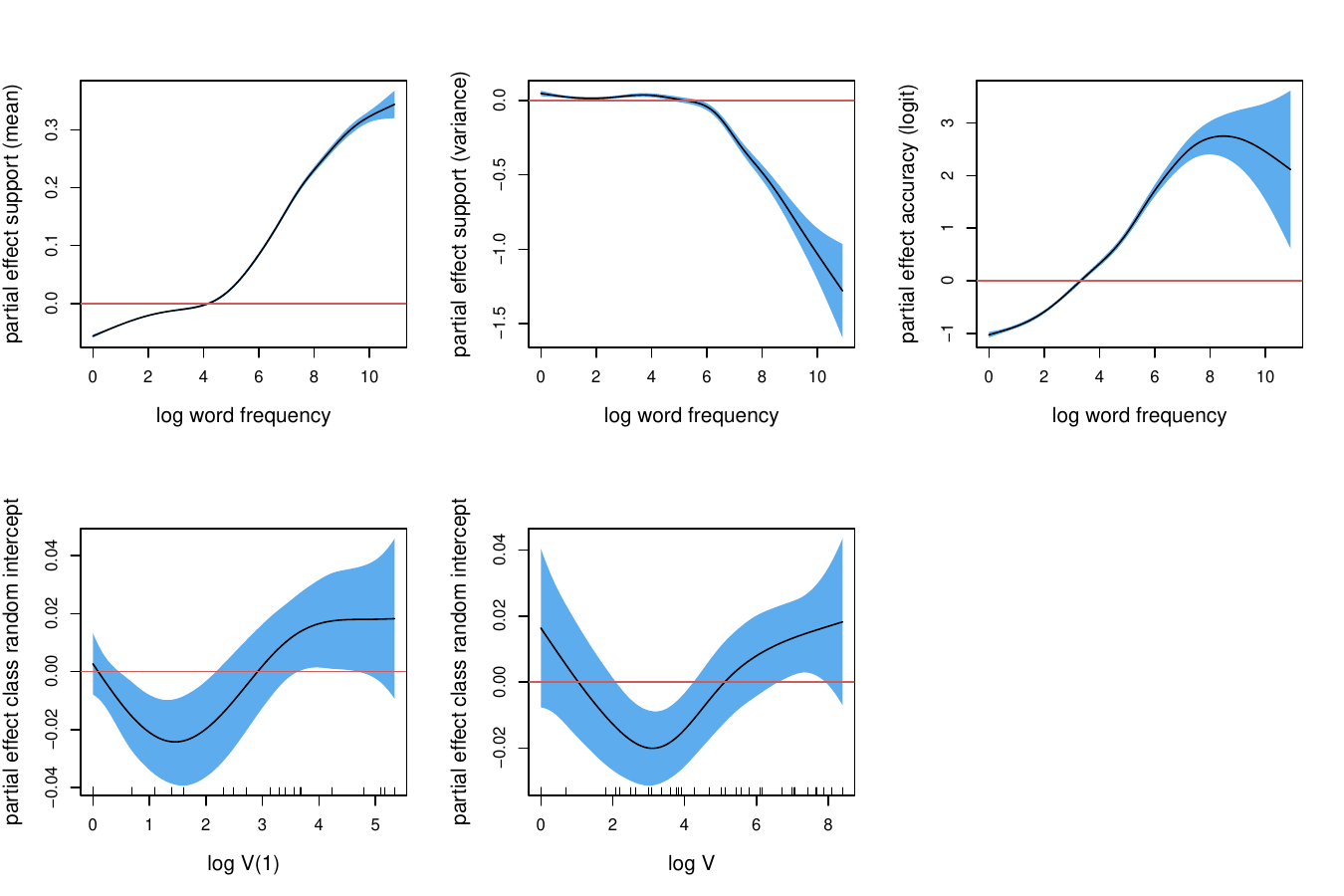}
    \caption{\textbf{Usage-based learning with FIL} for the \textbf{where-system}. The top left and center panels show the partial effects of frequency as predictor of support (left panel: mean, center panel: variance). The top right panel presents the partial effect in a logistic GAM predicting accuracy (all $p < 0.0001$). The lower panels clarify how log V(1) and log (V) predict the by-class random intercepts taken from the GAM fitted to the supports ($p = 0.0160$ for the TPRS smooth for log (V1), and $p = 0.0163$ for the TPRS smooth for log V).
    }
    \label{fig:FILprodWhere}
\end{figure}

\clearpage

Finally, we consider the consequences of a usage-based approach to the where-system. We calculated the support in the where system for a word's form, as well as the corresponding accuracy. {\color{black}A form is predicted accurately if and only if the support for that word form in the synthesis-by-analysis feedback loop is greater than the support for any other candidate form.}  We then fitted a Gaussian Location-Scale GAM to the support measures, and a logistic GAM to the accuracies, with as predictors log word frequency, case, number, and their interaction.  Including the interaction of case and number reduced the AIC by  890.7 units compared to a model with only main effects for case and number.  The upper panels of Figure~\ref{fig:FILprodWhere} present the partial effect of log frequency.  As expected, both support and accuracy increase with greater word frequencies.  

From the model fitted to the support measures, we extracted the random intercepts for inflectional class, and examined to what extent these random intercepts co-vary with measures of class productivity.  Modest non-linear effects were present only for log V(1) and for log (V). These partial effects are shown in the bottom panels of Figure~\ref{fig:FILprodWhere}.  For both measures, the trend is U-shaped.  For most of the range, the effect is in the expected direction, but for the lowest values, model performance as gauged with the support measure is better than expected for inflectional classes with very few types and hardly any hapax legomena.

In summary, modeling the what-system with a usage-based approach using FIL results, unsurprisingly, in a strong effect of learning on target correlations and accuracy.  The by-class random intercepts, the statistical signatures of by-class learning accuracy, are well-supported statistically ($p < 0.0001$), but do not co-vary systematically with productivity measures.  For the where-system, strong frequency effects also emerged.  By-class learning signatures (by class random intercepts) entered into rather weak non-linear relations with two of the productivity measures. All these effects arise in the mapping from form to meaning in the synthesis by analysis feedback loop. As a FIL-based comprehension mapping by itself shows reasonable correlations with productivity measures (see Figure~\ref{fig:comprehension_scatterplots}), we conclude that in a usage-based model, word frequency is the dominant factor for both the what-system and the where-system, with a very minor role (at best) for class productivity.   If the present computational models are on the right track \citep[but see][for an alternative approach]{Heitmeier:Chuang:Baayen:2024}, the productivity of inflectional classes plays a role primarily in comprehension. Given that comprehension precedes production in ontogeny, and that adult speakers understand many more words that they themselves produce, this result is perhaps unsurprising.  


\section{General discussion}

\textcolor{black}{Our findings demonstrate that the DLM can learn the complexities of Finnish nominal inflection without relying on pre-defined, explicit inflectional class features. This challenges the traditional assumption that mastering such abstract categories is necessary for learners to navigate complex morphology. The DLM's success appears rooted in its ability to leverage statistical regularities inherent in the mapping between form vectors and distributional semantic vectors. The model's ability to capture these regularities was significantly enhanced by using 4-gram input representations, which provide a more nuanced encoding of Finnish morphophonology, effectively capturing longer suffixes and stem changes characteristic of the language, compared to 3-grams (see Table \ref{tab:comp_acc}). Specifically, our Linear Discriminant Analyses revealed systematic semantic distinctions correlating with inflectional class membership within the embeddings, providing the structured information needed for the model to generalize effectively, particularly when case and number contexts were considered.}

\textcolor{black}{This conclusion--that learning can proceed effectively without mastery of complete abstract structures--aligns strongly with independent computational evidence from \cite{janda2021less}. Their work showed that a sequence-to-sequence model trained on realistic, frequency-ordered single word forms ultimately outperformed a model trained on full paradigms, suggesting that exposure to such complete paradigms (the traditional basis for defining inflectional classes) may be unnecessary or even counterproductive for robust generalization. Both our study using form-meaning mappings in the DLM and their study using sequence mappings point towards learning mechanisms, potentially mirroring those used by humans, that exploit statistical patterns and overlapping partial structures within usage data rather than requiring explicit representation of comprehensive linguistic categories like inflectional classes.}

\textcolor{black}{Our findings also resonate with other research employing computational cognitive modeling to probe the status of abstract grammatical categories. For instance, \cite{divjak2025learnability}, studying Polish aspect, found through NDL simulations and behavioral validation that models based on concrete usage patterns outperformed those relying on abstract semantic features, similarly questioning the operational necessity of high-level grammatical abstraction \citep[see also][]{romain2022makes,divjak2024nature}.}

\textcolor{black}{This perspective on learning inflection also prompts a re-evaluation of the distinction between rules and exceptions, often invoked in discussions of morphological productivity. Formal approaches like Yang's Tolerance Principle \citep{yang2016price}, for instance, propose a quantifiable threshold: productive rules are maintained only if the burden of exceptions remains manageable. Applying such a principle to Finnish nominal inflection, however, is problematic. If one were to designate a single, large inflectional class as the default 'rule', then nearly all other classes function inherently as structured sets of `exceptions', varying primarily in size and degree of deviation. The descriptive practice of establishing inflectional classes in Finnish itself involves grouping exceptions to a hypothetical broader pattern. Furthermore, our computational results complicate a strict rule-versus-rote-memorization dichotomy. Even for less productive classes, which under a Tolerance Principle view might be candidates for rote learning, the DLM demonstrates an ability to generalize and make correct predictions, albeit often less reliably than for highly productive classes. This suggests learnability extends across the spectrum of productivity, as evidenced by the varying levels of performance across classes (e.g., the by-class random intercepts for production models, discussed in relation to Fig. \ref{fig:deepWeave}). Productivity, therefore, appears less like a binary switch governed by exception counts and more like a gradient property emerging from 
a learning process that is driven by the statistical structure of the input.
This gradience poses a fundamental challenge to quantifying exceptions as required by the Tolerance Principle: if identifying a set of exceptions allows for the definition of a new sub-regularity or class, the process of counting becomes ill-defined. This inherent complexity suggests frameworks strictly separating rules from exceptions may be ill-suited for the intricate system of Finnish morphology, where learning necessarily involves mastering numerous idiosyncratic patterns (e.g., quantitative and qualitative consonant gradation) alongside broader generalizations.}

\textcolor{black}{The gradient nature of productivity suggested by our results finds resonance in usage-based network models, particularly the framework proposed by \cite{bybee1995regular}. While perhaps less specified for generating the precise quantitative predictions of computational models like the DLM or rule-based accounts like the Tolerance Principle, Bybee's model offers valuable insights into the role of token frequency, conceptualized as `lexical strength'. According to this view, words with high token frequency develop significant lexical strength, potentially becoming autonomous lexical units stored and accessed independently, rather than being generated via abstract rules or schemas each time they are encountered. This autonomy can have complex consequences for the learning and representation of morphology. On one hand, highly frequent, autonomous forms might be less reliant on, or even resistant to, participation in broader productive patterns, potentially contributing to the preservation of irregularities. On the other hand, the very existence of strong, frequent exemplars (especially regular ones) can serve to reinforce the schemas from which less frequent forms are generalized. This frequency-driven differentiation aligns well with the strong frequency effects observed in our DLM experiments, particularly the dominance of frequency over abstract productivity measures in the FIL production models (Fig. \ref{fig:prodFILwhat}). Bybee's perspective provides a useful qualitative lens, suggesting the mental lexicon is a dynamic system where individual item properties like frequency interact significantly with emergent generalizations, consistent with our finding that productivity appears gradient rather than categorical.}

\textcolor{black}{In conclusion, this study demonstrates that the Discriminative Lexicon Model can effectively model the complex nominal inflection of Finnish without recourse to explicit inflectional class features. This suggests that mastery of such intricate systems, by computational models and potentially by human learners, may rely less on acquiring abstract linguistic categories and more on extracting statistical regularities from the input---specifically, the correlations between form and meaning captured within distributional semantic spaces. Our findings, supported by the results of \cite{janda2021less} emphasizing learning from partial, frequency-weighted input, and consistent with usage-based principles regarding lexical strength and emergent generalization \citep{bybee1995regular}, point towards a view of morphological knowledge where gradient productivity and pattern sensitivity supersede rigid rule-exception dichotomies \citep[cf.][]{yang2016price}. \textcolor{black}{This perspective, emphasizing the extraction of statistical regularities from usage, gains further support from studies on other grammatical phenomena, such as Polish aspect, where computational models grounded in associative learning and validated by user behavior suggest that concrete contextual and lexical information are stronger predictors of choice than abstract grammatical categories \citep{divjak2025learnability}. Together, these findings bolster the view that complex grammatical systems can be mastered by leveraging sophisticated pattern detection over usage data, potentially bypassing the need for explicitly represented abstract structures like inflectional classes or invariant aspectual meanings.} While inflectional classes remain valuable descriptive tools with high educational value for L2 learners, our work indicates that the underlying cognitive or computational mechanisms might achieve generalization through sophisticated pattern detection operating over usage data, challenging long-held assumptions about the necessity of explicitly represented abstract structures for navigating morphological complexity.}

\bibliography{data,bibliography}

\begin{thebibliography}{}

\bibitem[Anshen and Aronoff, 1981]{Anshen:Aronoff:81}
Anshen, F. and Aronoff, M. (1981).
\newblock Morphological productivity and morphological transparency.
\newblock {\em The Canadian Journal of Linguistics}, 26:63--72.

\bibitem[Baayen, 1992]{Baayen:91:YoM}
Baayen, R.~H. (1992).
\newblock Quantitative aspects of morphological productivity.
\newblock In Booij, G.~E. and van Marle, J., editors, {\em Yearbook of Morphology 1991}, pages 109--149. Kluwer Academic Publishers, Dordrecht.

\bibitem[Baayen, 1993]{Baayen:92:YoM}
Baayen, R.~H. (1993).
\newblock On frequency, transparency, and productivity.
\newblock In Booij, G.~E. and van Marle, J., editors, {\em Yearbook of Morphology 1992}, pages 181--208. Kluwer Academic Publishers, Dordrecht.

\bibitem[Bojanowski et~al., 2017]{Bojanowski:2017:fastText}
Bojanowski, P., Grave, E., Joulin, A., and Mikolov, T. (2017).
\newblock {Enriching Word Vectors with Subword Information}.
\newblock {\em Transactions of the Association for Computational Linguistics}, 5:135--146.

\bibitem[Browman and Goldstein, 1992]{Browman:Goldstein:92}
Browman, C. and Goldstein, L. (1992).
\newblock Articulatory {P}honology: {A}n {O}verview.
\newblock {\em Phonetica}, 49:155--180.

\bibitem[Bybee, 1995]{bybee1995regular}
Bybee, J. (1995).
\newblock Regular morphology and the lexicon.
\newblock {\em Language and cognitive processes}, 10(5):425--455.

\bibitem[Divjak et~al., 2025]{divjak2025learnability}
Divjak, D., Milin, P., and Borowski, M. (2025).
\newblock On the learnability of aspectual usage.
\newblock {\em Corpus Linguistics and Linguistic Theory}.

\bibitem[Divjak et~al., 2024]{divjak2024nature}
Divjak, D., Testini, I., and Milin, P. (2024).
\newblock On the nature and organisation of morphological categories: verbal aspect through the lens of associative learning.
\newblock {\em Morphology}, 34(3):243--280.

\bibitem[Eronen, 1994]{eronen1994}
Eronen, R. (1994).
\newblock Taivutuksen osoittaminen perussanakirjassa.
\newblock {\em Kielikello}, 3.
\newblock Miten sanakirjaa luetaan?

\bibitem[Evans and Gazdar, 1996]{Evans:Gazdar:96}
Evans, R. and Gazdar, G. (1996).
\newblock {DATR}: A language for lexical knowledge.
\newblock {\em Computational {L}inguistics}, 22:167--216.

\bibitem[Fasiolo et~al., 2017]{Fasiolo:Goude:Nedellec:Wood:2017}
Fasiolo, M., Goude, Y., Nedellec, R., and Wood, S. (2017).
\newblock Fast calibrated additive quantile regression.
\newblock {https:\/\/github.com\/mfasiolo\/qgam}.

\bibitem[Good, 1953]{Good:1953}
Good, I.~J. (1953).
\newblock The population frequencies of species and the estimation of population parameters.
\newblock {\em Biometrika}, 40:237--264.

\bibitem[Grave et~al., 2018]{grave2018learning}
Grave, E., Bojanowski, P., Gupta, P., Joulin, A., and Mikolov, T. (2018).
\newblock Learning word vectors for 157 languages.
\newblock {\em arXiv preprint arXiv:1802.06893}.

\bibitem[Haarala, 1994]{haarala1990}
Haarala, R. (1990-1994).
\newblock {\em Suomen kielen perussanakirja}.
\newblock Edita Oyj, Helsinki.

\bibitem[Hakulinen et~al., 2004]{vilkuna2004iso}
Hakulinen, A., Vilkuna, M., Korhonen, R., Kovisto, V., Heinonen, T.~R., and Alho, I. (2004).
\newblock Iso suomen kielioppi.
\newblock In {\em SKS: n toimituksia 950}. Suomalaisen Kirjallisuuden Seura.

\bibitem[Heitmeier et~al., 2023]{Heitmeier:Chuang:Axen:Baayen:2023}
Heitmeier, M., Chuang, Y., Axen, S., and Baayen, R.~H. (2023).
\newblock Frequency-informed linear discriminative learning.
\newblock {\em Front. Hum. Neurosci., Sec. Speech and Language}, 17.

\bibitem[Heitmeier et~al., 2025]{Heitmeier:Chuang:Baayen:2024}
Heitmeier, M., Chuang, Y.-Y., and Baayen, R.~H. (2025).
\newblock {\em {The Discriminative Lexicon: {T}heory and implementation in the {J}ulia package JudiLing}}.
\newblock {Cambridge University Press}, Cambridge.
\newblock in press.

\bibitem[Hovdhaugen et~al., 2000]{hovdhaugen2000history}
Hovdhaugen, E., Karlsson, F., Henriksen, C.~C., and Sigurd, B. (2000).
\newblock {\em The history of linguistics in the Nordic countries}.
\newblock Citeseer.

\bibitem[Janda and Tyers, 2021]{janda2021less}
Janda, A.~L. and Tyers, M.~F. (2021).
\newblock Less is more: why all paradigms are defective, and why that is a good thing.
\newblock {\em Corpus linguistics and linguistic theory}, 17(1):109--141.

\bibitem[J\"{a}rvikivi, 2003]{Jarvikivi:2003}
J\"{a}rvikivi, J. (2003).
\newblock {\em Allomorphy and morphological salience in the mental lexicon}.
\newblock University of Joensuu, Joensuu.

\bibitem[J{\"a}rvikivi and Niemi, 2002a]{jarvikivi2002allomorphs}
J{\"a}rvikivi, J. and Niemi, J. (2002a).
\newblock Allomorphs as paradigm indices: On-line experiments with finnish free and bound stems.
\newblock {\em SKY journal of linguistics}, (15):119--143.

\bibitem[J{\"a}rvikivi and Niemi, 2002b]{jarvikivi2002form}
J{\"a}rvikivi, J. and Niemi, J. (2002b).
\newblock Form-based representation in the mental lexicon: Priming (with) bound stem allomorphs in finnish.
\newblock {\em Brain and language}, 81(1):412--423.

\bibitem[Karlsson, 1977]{karlsson1977eraista}
Karlsson, F. (1977).
\newblock Er{\"a}ist{\"a} morfologian teorian ajankohtaisista ongelmista.
\newblock {\em Sananjalka}, 19(1):26--56.

\bibitem[Karlsson, 1983]{karlsson1983suomen}
Karlsson, F. (1983).
\newblock Suomen kielen {\"a}{\"a}nne-ja muotorakenne [the phonological and morphological structure of finnish].
\newblock {\em Werner S{\"o}derstr{\"o}m, Juva}.

\bibitem[Karlsson, 1985]{karlsson1985paradigms}
Karlsson, F. (1985).
\newblock Paradigms and word forms.
\newblock {\em Studia gramatyczne}, 7:135--154.

\bibitem[Karlsson, 1986]{karlsson1986frequency}
Karlsson, F. (1986).
\newblock Frequency considerations in morphology.
\newblock {\em STUF-Language Typology and Universals}, 39(1-4):19--28.

\bibitem[Karlsson, 1997]{karlsson1997yleinen}
Karlsson, F. (1997).
\newblock {\em Yleinen kielitiede Suomessa kautta aikojen}.
\newblock Yleisen kielitieteen laitos, Helsingin yliopisto.

\bibitem[Karlsson, 1998]{karlsson1998vuosien}
Karlsson, F. (1998).
\newblock Vuosien 1906-1915 kielioppikomitean mietinn{\"o}n vastaanotto.
\newblock {\em Viritt{\"a}j{\"a}}, 102(1):2--2.

\bibitem[Karlsson, 2000]{karlsson2000setala}
Karlsson, F. (2000).
\newblock En set{\"a}l{\"a} vaarallisilla vesill{\"a}: Tieteellisen vallank{\"a}yt{\"o}n, k{\"a}ytt{\"a}ytymisen ja perinteen analyysi [en set{\"a}l{\"a} in dangerous waters: An analysis of scientific power, behaviour and tradition].
\newblock {\em Helsinki: Finnish Literature Society (SKS)}.

\bibitem[Lieberman et~al., 2007]{lieberman2007quantifying}
Lieberman, E., Michel, J.-B., Jackson, J., Tang, T., and Nowak, M.~A. (2007).
\newblock Quantifying the evolutionary dynamics of language.
\newblock {\em Nature}, 449(7163):713--716.

\bibitem[L\"onnrot, 1841]{lonnrot1841}
L\"onnrot, E. (1841).
\newblock Bidrag till finska språkets grammatik.
\newblock {\em Suomi}, 1(4):11--39.

\bibitem[Loo et~al., 2018a]{Loo:2018b}
Loo, K., Jaervikivi, J., and Baayen, R. (2018a).
\newblock Whole-word frequency and inflectional paradigm size facilitate {E}stonian case-inflected noun processing.
\newblock {\em Cognition}, 175:20--25.

\bibitem[Loo et~al., 2018b]{Loo:2018a}
Loo, K., Jaervikivi, J., Tomaschek, F., Tucker, B., and Baayen, R. (2018b).
\newblock Production of {E}stonian case-inflected nouns shows whole-word frequency and paradigmatic effects.
\newblock {\em Morphology}, 1(28):71--97.

\bibitem[Niemi et~al., 1994]{Niemi.etal:94}
Niemi, J., Laine, M., and Tuominen, J. (1994).
\newblock Cognitive morphology in {F}innish: foundations of a new model.
\newblock {\em Language and {C}ognitive {P}rocesses}, 9:423--446.

\bibitem[Nikolaev et~al., 2022]{nikolaev2022generating}
Nikolaev, A., Chuang, Y.-Y., and Baayen, R.~H. (2022).
\newblock A generating model for finnish nominal inflection using distributional semantics.
\newblock {\em The Mental Lexicon}, 17(3):368--394.

\bibitem[Nikolaev and Niemi, 2008]{nikolaev2008suomen}
Nikolaev, A. and Niemi, J. (2008).
\newblock Suomen nominien taivutusj{\"a}rjestelm{\"a}n produktiivisuuden indekseist{\"a}.
\newblock {\em Viritt{\"a}j{\"a}}, 112(4):518--544.

\bibitem[Orzechowska and Baayen, 2025]{Orzechowska:Baayen:2025}
Orzechowska, P. and Baayen, R.~H. (2025).
\newblock Polish phonology and morphology through the lens of distributional semantics.
\newblock Manuscript, University of Poznan, University of Tuebingen.

\bibitem[Petræus, 1968]{petraeus19681649}
Petræus, A. (1968).
\newblock 1649: Linguae finnicae brevis institutio.
\newblock {\em Aboae. SWK}, 1:1649--1816.

\bibitem[Raun, 1959]{raun1959suomen}
Raun, A. (1959).
\newblock Suomen kielen deklinaatioista ja konjugaatioista.
\newblock {\em Viritt{\"a}j{\"a}}, 63(3):348--348.

\bibitem[Roark and Sproat, 2007]{roark2007computational}
Roark, B. and Sproat, R. (2007).
\newblock {\em Computational approaches to morphology and syntax}, volume~4.
\newblock OUP Oxford.

\bibitem[Romain et~al., 2022]{romain2022makes}
Romain, L., Ez-zizi, A., Milin, P., and Divjak, D. (2022).
\newblock What makes the past perfect and the future progressive? experiential coordinates for a learnable, context-based model of tense and aspect.
\newblock {\em Cognitive Linguistics}, 33(2):251--289.

\bibitem[Sadeniemi and Vesikansa, 1961]{nykysuomen_sanakirja}
Sadeniemi, M. and Vesikansa, J. (1951-1961).
\newblock {\em Modern Finnish Dictionary}.
\newblock WSOY, Helsinki.

\bibitem[Schot-Saikku, 1994]{schot1994taivutuksen}
Schot-Saikku, P. (1994).
\newblock Taivutuksen mallintamisesta ulkomaalaisopetuksessa.
\newblock {\em Viritt{\"a}j{\"a}}, 98(2):248--248.

\bibitem[Set{\"a}l{\"a}, 1898]{setala1898suomen}
Set{\"a}l{\"a}, E.~N. (1898).
\newblock {\em Suomen Kielioppi {\"a}{\"a}nne-jạ sanaoppi oppikoulua ja omin p{\"a}in opiskelua varten Kirjoittanut Emil Nestor Set{\"a}l{\"a}}.
\newblock Kustannusosakeyhti{\"o} Otava.

\bibitem[Snell, 2024]{snell2024pong}
Snell, J. (2024).
\newblock Pong: A computational model of visual word recognition through bihemispheric activation.
\newblock {\em Psychological Review}.

\bibitem[Vhael, 1968]{vhael19681733}
Vhael, B. (1968).
\newblock 1733: Grammatica fennica.
\newblock {\em Aboae. SWK}, 1:1649--1816.

\bibitem[Vihonen, 1978a]{vihonen1978}
Vihonen, S. (1978a).
\newblock Suomen kielen oppikirja 1600-luvulla.
\newblock {\em Studia Philologica Jyväskyläensia 11}.

\bibitem[Vihonen, 1978b]{vihonen1978suomen}
Vihonen, S. (1978b).
\newblock Suomen varhaiskieliopit ja 1900-luvun kielentutkimus.
\newblock {\em Viritt{\"a}j{\"a}}, 82(2):160--160.

\bibitem[Wiik, 1967]{wiik1967suomen}
Wiik, K. (1967).
\newblock {\em Suomen kielen morfofonemiikkaa: yritys soveltaa transformaatioteoriaa suomen yleiskielen sanojen taivutukseen}.
\newblock Phonetics Department of the University of Turku.

\bibitem[Wiik, 1989]{wiik1989suomen}
Wiik, K. (1989).
\newblock {\em Suomen kielen morfofonologian historia: Nominien taivutus 1649-1820}.
\newblock Turun yliopisto, fonetiikka.

\bibitem[Wikstr{\"o}m, 1832]{wikstrom1832forsok}
Wikstr{\"o}m, M. (1832).
\newblock F{\"o}rs{\"o}k till en finsk grammatita, framst{\"a}llande en enda dekli-nation och en enda conjugation.
\newblock {\em Wasa. SWK}, 2:1818--1832.

\bibitem[Williams et~al., 2019]{williams2019quantifying}
Williams, A., Cotterell, R., Wolf-Sonkin, L., Blasi, D., and Wallach, H. (2019).
\newblock Quantifying the semantic core of gender systems.
\newblock {\em arXiv preprint arXiv:1910.13497}.

\bibitem[Wood, 2017]{Wood:2017}
Wood, S.~N. (2017).
\newblock {\em {Generalized Additive Models}}.
\newblock Chapman \& {H}all/{CRC}, New {Y}ork.

\bibitem[Yang, 2016]{yang2016price}
Yang, C. (2016).
\newblock {\em The price of linguistic productivity: How children learn to break the rules of language}.
\newblock MIT press.

\end{thebibliography}

\end{document}